\documentclass{egpubl}
\usepackage{sca2020}

\SpecialIssuePaper         %

\usepackage[T1]{fontenc}
\usepackage{dfadobe}  
\usepackage{amsfonts}
\usepackage{cite}  %
\usepackage{caption}
\usepackage{subcaption}
\BibtexOrBiblatex
\electronicVersion
\PrintedOrElectronic
\ifpdf \usepackage[pdftex]{graphicx} \pdfcompresslevel=9
\else \usepackage[dvips]{graphicx} \fi

\usepackage{egweblnk} 
\usepackage[dvipsnames]{xcolor}
\usepackage{amsmath}
\usepackage{bm}
\usepackage[normalem]{ulem}

\newcommand{\new}[1]{{{#1}}}

\newcommand{\overbarcal}[1]{\mkern 1.5mu\overline{\mkern-4.5mu#1\mkern-1.5mu}\mkern 1.5mu}
\newcommand{\overbarcalT}[1]{\mkern 1.5mu\overline{\mkern-0.25mu#1\mkern-0.5mu}\mkern 1.5mu}

\title[Fully Convolutional Graph Neural Networks for Parametric Virtual Try-On]
{Fully Convolutional Graph Neural Networks \\for Parametric Virtual Try-On}

\author[R. Vidaurre, I. Santesteban, E. Garces,  \& D. Casas]
{\parbox{\textwidth}{\centering 
		Raquel Vidaurre$^1$~~~~~~~~~~~~~~
		Igor Santesteban$^1$~~~~~~~~~~~~~~   
		Elena Garces$^2$~~~~~~~~~~~~~~
		Dan Casas$^1$
	}
	\\
	{\parbox{\textwidth}{\centering $^1$Universidad Rey Juan Carlos, Madrid, Spain.\\
		$^2$ SEDDI Labs, Madrid, Spain.}
	}
}

\begin{document}
	
	\teaser{
\includegraphics[width=\linewidth,trim={0 12pt 0 195pt},clip]{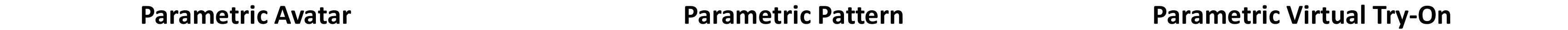}
\includegraphics[width=\linewidth,trim={0 40pt 0 20pt},clip]{figures/teaser5.pdf}
		\centering
		\caption{Our method predicts the 3D draping for an arbitrary body shape and garment parameters at interactive rates. From left to right, a variety of body shapes obtained from a parametric avatar model, different \new{2D panel} configurations of our paremeterized garment types, and corresponding dressed \new{3D} bodies generated with our novel fully convolutional approach. }
		\label{fig:teaser}
	}
	\maketitle
	\begin{abstract}
		We present a learning-based approach for virtual try-on applications based on a fully convolutional graph neural network.
		In contrast to existing data-driven models, which are trained for a specific garment or mesh topology, our fully convolutional model can cope with a large family of garments, \new{represented as parametric predefined 2D panels with arbitrary mesh topology}, including long dresses, shirts, and tight tops. 
		Under the hood, our novel geometric deep learning approach learns to drape 3D garments by
		decoupling the three different sources of deformations that condition the fit of clothing: garment type, target body shape, and material.
		Specifically, we first learn a regressor that predicts the 3D drape of the input parametric garment when worn by a mean body shape. 
		Then, after a mesh topology optimization step where we generate a sufficient level of detail for the input garment type, we further deform the mesh to reproduce deformations caused by the target body shape.
		Finally, we predict fine-scale details such as wrinkles that depend mostly on the garment material.
		We qualitatively and quantitatively demonstrate that our fully convolutional approach outperforms existing methods in terms of generalization capabilities and memory requirements, and therefore it opens the door to more general learning-based models for virtual try-on applications.  

		
	\end{abstract}

\section{Introduction}
\label{sec:introduction}
The digitization of clothing is a long-standing goal in Computer Graphics and Animation, with important applications in many areas including garment design, virtual try-on, film visual effects, and video games.
The classic --and still nowadays prevalent-- approach to tackle the digitization of clothing 
is based on physics-based methods that simulate the deformations of garments in contact with the body~\cite{nealen2006physically,narain2012arcsim,cirio2014yarn}. 
Despite the tremendous realism achieved with these methods, their high computational cost and potential instabilities hinder their deployment in everyday applications.
Many methods have been proposed to limit such computational cost, including coarse discretizations of the cloth \cite{kavan2011upsampling,zurdo2012animating}, simplified simulation models \cite{muller2010wrinkle}, subspaces \cite{de2010stable,fulton2019latent}, and tailored GPU-based solvers \cite{tang2018gpu}. 

As an alternative to physics-based methods, data-driven solutions aim at learning a function that approximates the ground truth deformations. 
Initial attempts used linear models \cite{guan2012drape}, which struggle to reproduce the complexity of wrinkles and dynamics inherent in clothing.
More recently, deep learning methods \cite{wang2018multimodalspace,santesteban2019virtualtryon,patel2020tailor,ma2020dressing3d} have demonstrated that it is possible to learn realistic and highly efficient models for cloth animation.
Most of these methods leverage existing parametric human models \cite{loper2015smpl} to use as input the shape and/or pose body parameters and output the corresponding deformed garment.
Despite the realism of the predicted animations, a common underlying limitation of existing data-driven methods is the dramatic inability to generalize to new garments and mesh topologies. 
While physics-based methods naturally generalize to almost arbitrary combinations of garment meshes, most data-driven solutions are trained for a specific garment or mesh topology.

The reason for such limitation are the architectures used in existing solutions, which typically comprise one or more fully-connected layers (also know as multilayer perceptrons, MLP)~\cite{wang2018multimodalspace,santesteban2019virtualtryon}. 
While this type of layers are known to be easy to use in any domain (images, meshes, etc.) since input values (\textit{e.g.,} pixels, vertices, etc.) are just flattened into a vector, they carry a number of limitations.
First, it constrains the size of the input vector to a fixed number, which enforces input meshes to have always the same topology. 
Second, it disregards spatial information due to the flattening vectors as input, which causes the loss of important local or neighboring information.
And third, it requires a large number of parameters, since all input nodes (or neurons) are densely connected to each other. 

In this work, we address such fundamental limitation in data-driven cloth by using a \textit{fully convolutional graph neural network} (FCGNN).
Fully convolutional architectures have shown to be successful in Euclidean domains, where the input data is regularly distributed over 2D or 3D grid. For example in image segmentation methods that take as input an image or volume of arbitrary size and produce a correspondingly-sized output with efficient per-pixel inference~\cite{shelhamer2017fully,milletari2016v}.
For non-Euclidian domains such as 3D meshes, the use of fully convolutional architectures is more challenging due to the non-trivial definition of a convolution operators in such irregular domain.
To this end, we leverage recent research that formally defines the required operators for graph-like structures \cite{defferrard2016convolutional,bronstein2017geometric} and propose a FCGNN that, given a \textit{3D parametric garment \new{(represented with known 2D panels)} with arbitrary mesh topology} (\textit{i.e.}, random) and a target body shape, outputs the accurate 3D draped garment.

Under the hood, our novel geometric deep learning approach learns to drape 3D garments by
decoupling the three different sources of deformations that condition the fit of %
clothing: garment type, target body shape, and material.
To this end, we initially build a parametric space for garment design that is capable of representing a large number of garment types, including loose dresses, shirts, t-shirts, and tight tops.
Using this design space, we create a dataset of 3D garments and use physics-based simulation \cite{narain2012arcsim} to dress a wide range of body shapes. We use this data to train three different networks, each of which serves for a different purpose in the virtual try-on pipeline. First, we learn a regressor that efficiently predicts, given the garment parameters, the coarse 3D drape of a garment onto the \textit{mean} body shape. 
Then, to provide sufficient surface detail to each garment type, we apply a mesh topology optimization step that generates a regular and homogeneous size triangular mesh.
Deformations caused by target body shape are modeled in our second deformation step, which consists of a regressor that deforms the topology-optimized \textit{mean shape} fitted garment as a function of the body shape.
Our final step further deforms the garment to account for material-specific deformations, which mostly produce fine-scale wrinkles.
Furthermore, we fine-tune our regressors with a novel \textit{self-supervised} ({i.e.}, does not require on physically-based simulated data) strategy that penalizes body-cloth collisions. 

All in all, our main contribution is a novel geometric deep learning framework that is able to cope with parametric garments \new{represented as predefined 2D panels}, arbitrary \new{mesh} topology, and any target body shape.
We discuss and evaluate the advantages of the proposed architecture, and compare with existing methods and other baselines. 
To the best of our knowledge, our approach is the first fully convolutional method (\textit{i.e.}, no fully connected layers are used) for data-driven cloth.

	\section{Related Work}
\label{sec:related_work}
The modeling of clothing has been approached in different ways. Here we discuss existing works by grouping them into simulation, 3D reconstruction, and data-driven models.
\paragraph*{Cloth Simulation.}
Physics-based simulation methods use discretizations of classical mechanics to deform cloth by solving an ordinary differential equation\cite{nealen2006physically}.
Based on this strategy many approaches have been proposed, with differences in the underlying representation, numerical solution methods, collision detection, and constraints. 
Despite the high level of realism shown with these approaches, capable of modeling even yarn mechanics~\cite{kaldor2010efficient,cirio2015efficient}, they usually have a significant runtime computational costs that hinders the use in interactive applications.

A wide variety of attempts exist to limit the computational cost. For example, using position-based dynamics~\cite{mueller2007pbd,kim2012long,muller2014strain}, which produce approximated but plausible results, but may lack the realism needed for real-world applications such as virtual try-on.
Other methods use subspaces or model reduction techniques \cite{de2010stable,sifakis2012fem,holden2019neuralphysics,fulton2019latent} to perform simulation in a reduced space. Projecting the equations of motion into the subspace is simple, but adding constraints or problem-specific details is challenging.
Alternatively, some methods speed up physics-based simulation by adding details to low-resolution simulated meshes~\cite{kavan2011upsampling,zurdo2012animating,gillette2015realtime} or simplified physical models~\cite{rohmer2010animation,muller2010wrinkle}.

Physics-based approaches have also been proposed from a design perspective, where the users specifies garment parameters and simulation is used to compute the 3D drape~\cite{umetani2011sensitive,berthouzoz2013parsing}.
We also predict the 3D drape given garment parameters, but we seek to skip the computationally-expensive simulation step while being capable of handling any garment designs, mesh topology, and target body shape.
Very recently, physic-based differentiable methods have been proposed
\cite{liang2019differentiable}, which also enable the efficient optimization of design parameters to produce the desired 3D drape.

\paragraph*{Cloth Reconstruction.}
As an alternative to simulation, 3D reconstruction methods aim at recovering the surface of real clothing. Reconstructed garments can potentially be used later to dress new subjects or train data-driven models, as we discuss in the next subsection.

Early attempts required customized clothing or special patterns to capture the deforming surface of a worn garment \cite{scholz2005garment,white2007capturing}.
Subsequent research by Bradley \textit{et al.}~\cite{bradley2008markerless} succeed at reconstructing sequences of markerless garments, with relatively low wrinkle level detail, using a multi-view stereo approach. Notably, their method even achieves temporally-coherent geometry by enforcing as-isometric-as-possible inter-frame constraints.
Detail upsampling techniques, similar to those used to enhance low-resolution simulated meshes, can be used in this context too \cite{popa2009wrinkling}.
Zhou \textit{et al.}~\cite{zhou2013garment} reduced the input requirements and reconstruct garments from a single image. They automatically detect the garment outline to roughly create a smooth 3D mesh that is further refined with shape-from-shading cues. 
A common limitation of these garment capture methods is their inability or significant difficulty to manipulate the reconstructed clothing, for example, as a function of the underlying body shape for virtual try-on applications.

A different trend in 3D reconstruction, usually referred to as \textit{performance capture}, aims at recovering the full body of a dressed actor while moving~\cite{starck2007surface,de2008performance,vlasic2008articulated,xu2018monoperfcap}.
Pioneering methods use a full-body 3D template of the actor that it is deformed using an optimization scheme such that it matches images captured using a multi-camera studio \cite{de2008performance,vlasic2008articulated}. Follow up methods reduced the input requirements by employing just a single depth camera~\cite{zhang2014dynamichuman,bogo2015detailed}, or even a monocular video \cite{xu2018monoperfcap,yang2018garmentrecovery}. Recently, template-based reconstruction methods have also been proposed for outdoor settings ~\cite{robertini2016model,xu2018monoperfcap}, and for animals on the wild~\cite{zuffi2018lions}.
Alternatively, template-free methods~\cite{starck2007surface} combine visual hull-based techniques with stereo reconstruction to extract per-frame surface of the actor. 
Even if re-animation of captured performances is possible~\cite{casas2014video4d,prada2016motiongraph}, the main limitation of these approaches is the single mesh output used to represent both the human body and clothing, which hinders the digital manipulation of the captured garment.

There exist works that address the problem of segmenting reconstructed 3D meshes into body and clothing layers.
Neophytou and Hilton~\cite{neophytou2014layered} estimate the underlying body shape by fitting a parametric human model, and learn a clothing deformation model with the residual of the fit. The learned model can be then used to dress different body shapes.
Similarly, Pons-Moll \textit{et al.}~\cite{pons2017clothcap} present a remarkable multi-camera system that is capable of reconstructing the underlying human shape and multiple garment layers, with fine wrinkle detail, at 60fps. Reconstructed garments can be transferred to new body shapes, but the dynamics in new sequences may look unrealistic since they are just a copy of the captured deformations.
Yang \textit{et al.}~\cite{yang2018analyzing} go one step beyond and enrich the captured dataset with simulated data that exhibits variations in clothing size and physical materials. This enables the learning of a richer garment deformation model, capable of representing semantic parameters such as material properties. 

Deep learning techniques have also been proposed to address the \textit{3D reconstruction} of garments. These methods circumvent the need for the error-prone model-fitting or optimization step in previous methods, and achieve a faster performance.
Dan{\v{e}}{\v{r}}ek \textit{et al.}~ \cite{danvevrek2017deepgarment} use synthetic data to train a CNN that regresses 3D vertex offsets to reconstruct a single garment from images. They require a known 3D template, and a tight crop of the image.
Alldieck \textit{et al.}~\cite{alldieck19cvpr} and Bhatnagar \textit{et al.}~\cite{bhatnagar2019multi} learn to reconstruct clothing and hair from video as displacements on top of SMPL human model \cite{loper2015smpl}.
DeepWrinkles \cite{lahner2018deepwrinkles} learns a 3D clothing deformation model from scans that is subsequently used to regress pose-dependent deformations for a specific garment.
Even if our approach is not aimed for reconstructing tasks, we also use deep learning to regress 3D garment deformations. In contrast to these methods, our model generalizes parametric garments, does not rely on \textit{a priori} known topology, and generalizes to different body shapes.
\paragraph*{Data-driven 3D Models.}
Inspired by the success of the extensive literature in statistical 3D human body models learned from scans 
\cite{anguelov2005scape,feng2015avatar,loper2015smpl,cheng2018parametric}, with recent works even capable of learning highly dynamic soft skin deformations \cite{santesteban2020softsmpl},
many methods have proposed to learn 3D clothing models from data. 
Guan~\textit{et al.}\cite{guan2012drape} use simulated data to lean to deform a garment as a function of the shape and pose of the underlying body. However, they rely on a linear model that struggles to learn fine details.
Similarly, Xu \textit{et al.}~\cite{xu2014sensitivity} retrieve garment parts from a simulated dataset to synthesize pose-dependent clothing meshes, but shape deformations are not modeled.  
More recent methods use machine learning to predict garment deformations as a function of body pose alone ~\cite{gundogdu2019garnet}, or pose and shape~\cite{santesteban2019virtualtryon}, some even capable of learning style \cite{patel2020tailor} or animation dynamics as a function of fabric parameters~\cite{wang2019intrisicspace}.
A common limitation of these methods is the need to train a regressor for each garment, which hinders their deployment to massive use.
In contrast, our method is able to learn to deform a large variety of garments using the same model.

The \textit{design} of garments have been also tacked with data-driven models.
Particularly relevant for us is the work of Wang \textit{et al}~\cite{wang2018multimodalspace}, who learn a multi-modal subspace that allows editing the design of a specific garment using both 2D panel size and a sketch of the desired drape. Given a target body shape, the method outputs the 3D draped garment according to different input modalities.
In contrast, we focus on the virtual try-on scenario instead of the design aspect. The fully-convolutional machinery of our method is able to cope with a wider range of clothing variability, ranging from tight tops to long loose dresses, using the \textit{same} trained model.

\begin{figure*}
	\centering
	\includegraphics[width=0.93\linewidth]{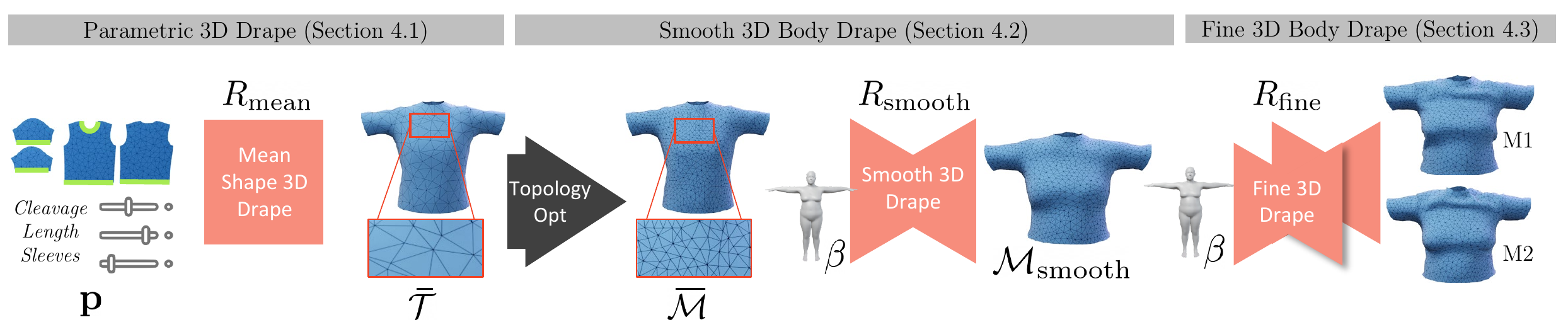}
	\caption{Our pipeline for parametric virtual try-on. First, we estimate a parametric 3D mesh $\overbarcalT{\mathcal{T}}$ for the \emph{mean} human shape given a set of design parameters $\mathbf{p}$. Then, after a topology optimization step that allows us to deal with any input topology and fine geometric details, a fully convolutional regressor {\normalfont $R_{\text{smooth}}$} estimates smooth 3D deformations {\normalfont $\mathcal{M}_{\text{smooth}}$} for a \emph{target} human shape $\beta$. Finally, this mesh is further refined using another fully convolutional regressor {\normalfont $R_{\text{fine}}$} to take into account fine wrinkles and material-specific deformations, represented as M1 and M2 in the figure.}
	\label{fig:pipeline}
\end{figure*}
Recently, following the success of graph CNNs (GCNNs) in non-euclidean domains such as triangular surfaces~\cite{defferrard2016convolutional,bronstein2017geometric}, some methods have explored the use of  GCNN to learn 3D mesh deformations from data. 
Tan \textit{et al.}~\cite{tan2018variational} propose a mesh variational autoencoder (VAE) to learn an efficient latent space for 3D surfaces. The generative nature of the subspace enables  to sample, interpolate, and explore deformable shapes.
In a similar spirit, Ranjan \textit{et al.}~\cite{ranjan2018generating} use a graph convolutional autoencoder to learn a generative model for 3D faces.
Closer to ours are the works that use GCNNs to learn garment deformations.
Ma \textit{et al.}~\cite{ma2020dressing3d} employ a conditional mesh VAE-GAN~\cite{larsen2016vaegan} to compute additive displacements that are applied onto a full-body human mesh. Despite being able to reproduce global and local cloth deformations, they require a fully-connected layer and a fix size input mesh.
Similarly, Bertiche \textit{et al.} \cite{bertiche2020cloth3d} propose a model that is able to learn deformations for a large number of garments, but cannot cope with varying topology. 	 
In contrast, our method is fully convolutional and, \new{assuming a known garment design space with a predefined parametric 2D panel layout (see Figure \ref{fig:pipeline} left)}, we are able to cope with \textit{any} mesh topology with arbitrary number of vertices.

	\section{Overview}
\label{sec:overview}
Our goal is to predict the accurate 3D draping of garments, worn by any body shape, for virtual try-on purposes.
We put special emphasis on the ability to cope with a large variety of garments, a feature mostly ignored by existing works since it requires a model that can deal with varying topology input.
To this end, we propose the three-stage approach depicted in Figure \ref{fig:pipeline} that effectively decouples (and therefore, eases learning tasks) the different sources of deformations (\textit{e.g.,} due to garment type, body shape, or material) that condition the fit of clothing.

Following the traditional garment design workflow, our first step (Section \ref{sec:mean_drape}) uses a set of parameters $\mathbf{p}$ to define the 2D sewing patterns of a garment (\textit{e.g.}, sleeve length, chest circumference, etc.), and learns a regressor  $R_{\text{mean}}(\mathbf{p})=\overbarcalT{\mathcal{T}}$ to estimate the corresponding 3D mesh $\overbarcalT{\mathcal{T}}$ draped into the \textit{mean} human shape.
Then, in order to accurately represent all garments potentially designed with the parameters $\mathbf{p}$ (\textit{e.g.}, from tight sleeveless tops to loose dresses), our second step (Section \ref{sec:topology_opt}) computes an optimized mesh topology $\overbarcal{\mathcal{M}}$ from the regressed 3D drape $\overbarcalT{\mathcal{T}}$, and learns a regressor $R_{\text{smooth}}$
to predict a smooth (\textit{i.e.,} lacking fine wrinkles) fit  $\mathcal{M}_{\text{smooth}}$ onto the target body shape $\mathbf{\beta}$.
Lastly, the third step (Section \ref{sec:accurate_drape}) learns a regressor $R_{\text{fine}}$ %
that predicts a deformed mesh $\mathcal{M}_{\text{fine}}$ with the realistic draping of the garment into the \textit{target} body shape $\beta$.
Importantly, regressors $R_{\text{smooth}}$ and $R_{\text{fine}}$ are implemented in a novel fully convolutional graph neural network (FCGNN) that is able to cope with \textit{any} combination of garment, topology, and target body.
Furthermore, in order to resolve potential body-garment collisions due to small inaccuracies when predicting surface deformations (a common issue in learning-based garment deformation methods, \textit{e.g.} \cite{wang2018multimodalspace,santesteban2019virtualtryon}), in Section \ref{sec:selfsupervised_collisions} we propose a novel self-supervised strategy to fine-tune the regressor $R_{\text{fine}}$.
The proposed supervision is based on a geometric definition of the distance between garment vertices and body faces, and does not require ground truth data (\textit{i.e.}, avoids the need for expensive cloth simulation).
	\newcommand{\lowpmean}{\overbarcalT{\mathcal{T}}}
\newcommand{\highpmean}{\overbarcalT{\mathcal{M}}}
\newcommand{\highpbetasmooth}{\mathcal{M}_{\text{smooth}}}
\newcommand{\highpbetafine}{\mathcal{M}_{\text{fine}}}

\section{Garment Parametric Virtual Try-On}
\subsection{Parametric 3D Drape}
\label{sec:mean_drape}
In order to predict the 3D draping of garments for virtual try-on applications, we first need to define the actual garment type.
Inspired by the traditional clothing manufacturing workflow, and similar to existing works \cite{umetani2011sensitive,wang2018multimodalspace}, we characterize garment design properties using 2D sewing patterns.
However, our observation is that we can use a single 2D layout to model a large family of garments by simply editing the length of some specific parts.
For example, a tight sleeveless top and a long dress can be represented with the same 2D layout contour, with differences just in terms of size of each layout part.
This is in contrast to the common use of 2D layouts, which are usually edited only to model size or small style changes.
For example, Wang \textit{et al.}~\cite{wang2018multimodalspace} only allow minimal edits in 2D to change the style of the garment, and require independent models for dresses and t-shirts.

Based on this observation, our first step is learning to predict a \textit{coarse} 3D drape of a garment given a specific 2D sewing pattern. In particular, we encode the parameters of the 2D layout (\textit{e.g.}, sleeve length, chest circumference, etc.) in a vector $\mathbf{p}$ that is fed into a non-linear regressor
\begin{equation}
	R_{\text{mean}}(\mathbf{p})=\lowpmean
\end{equation}
that outputs the drape of the garment onto a \textit{mean} human shape (note that in the rest of the paper we will use the overline symbol to refer to mean-shape-related variables).
The motivation of this initial step is twofold: 
first, it roughly fits the garment on a generic human subject, which we use later in Section \ref{sec:topology_opt} to 
parameterize garment vertices using their closest body skinning weights; %
and second, it allows us to disentangle garment type-dependent deformations (\textit{i.e.,} that depend on $\mathbf{p}$) from material-dependent and body shape-dependent deformations.

\begin{figure}
	\centering
	\includegraphics[width=1.0\columnwidth]{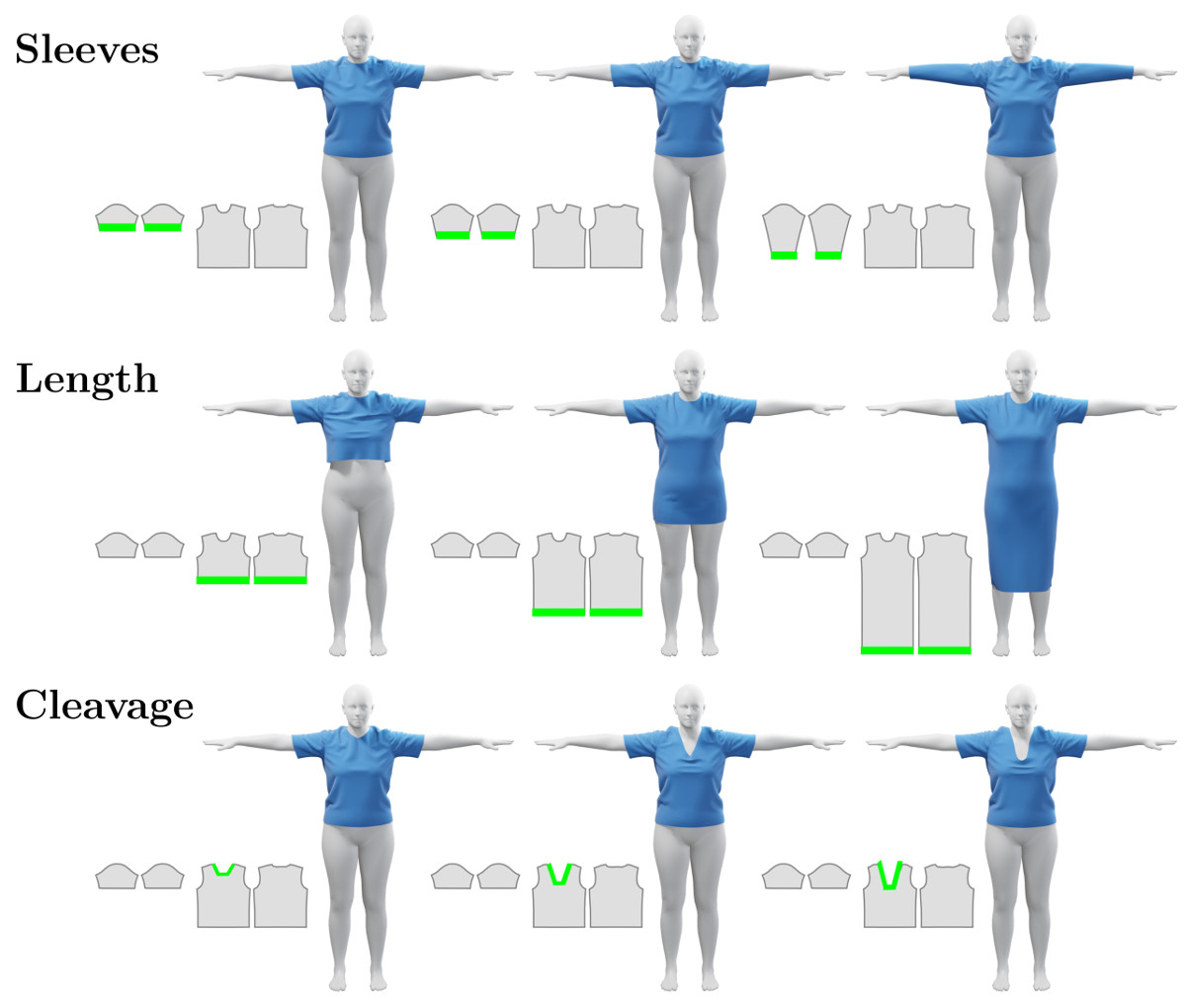}
	\caption{Sewing pattern parameters (rows) used to build our dataset of garments. Each column shows the effect of the minimum, mean, and maximum values for each parameter.}
	\label{fig:sewing_patterns}
\end{figure}

To train our regressor $R_{\text{mean}}(\mathbf{p})$ we build a dataset of 3D garments by manipulating a single 2D layout. 
Specifically, as shown in Figure \ref{fig:sewing_patterns}, we manually edit parts of the 2D panels to design a family of garments including tops, t-shirts, sweaters, and short and long dresses. We then label each sample according to a set of measurements $\mathbf{p}$ in the corresponding 2D representation, and simulate the sample worn by a mean human shape using %
a state-of-the-art physics-based cloth simulator \cite{narain2012arcsim}, \new{with remeshing option turned off}, until it reaches equilibrium to obtain a 3D mesh $\lowpmean$ of the draped garment.
We implement the regressor $R_{\text{mean}}\colon \mathbb{R}^P \to \mathbb{R}^{3 \times {V^{\lowpmean}}}$ using a fully connected neural network that outputs the vertices positions of the mesh $\lowpmean$
with a predefined topology.

\subsection{Smooth 3D Body Drape}
\label{sec:smooth_drape}
\paragraph*{Mesh Topology Optimization.}
To accurately represent the draping of 3D garments with fine-scale detail it is necessary to use a topology with sufficient resolution (\textit{i.e.}, number of triangles) for each garment type. 
Since one of our goals is to build a model \new{that} can predict the deformations for a large family of garments, we need to adapt the topology of the mesh $\lowpmean$ depending on the type of garment.
To give a more practical example, we assume that the number of triangles required to represent high-quality draping of a t-shirt is smaller than those required for a long dress.

\begin{figure}
	\centering
	\includegraphics[width=1.0\columnwidth]{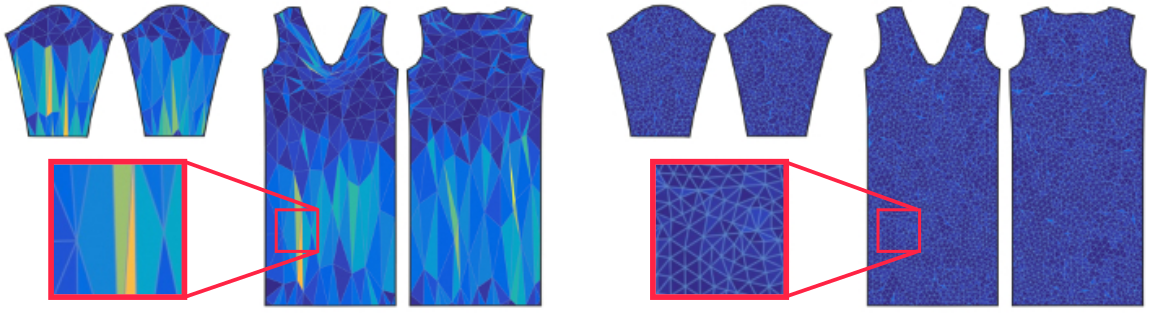}
	\caption{Garment type-dependent topology optimization, \new{here shown in 2D panel space}. \new{Left}: the 2D layout of a long dress design, with the template topology $\lowpmean$. \new{Right}: the same design after the topology optimization step, resulting in the mesh $\overbarcal{\mathcal{M}}$ with homogeneous triangle size and without degenerated geometry. 
	}
	\label{fig:retopo}
\end{figure}

\label{sec:topology_opt}
We model such garment type-dependent topology requirement by applying a remeshing operation to the coarse mean draped garment $\lowpmean$. 
Specifically, we generate a new mesh
\begin{equation}
\overbarcal{\mathcal{M}} = \phi(\lowpmean, \mathbf{p}, T_{\text{{dist}}}, T_{\text{{area}}}),
\end{equation}
where $\phi()$ is a remeshing operation that, given an input mesh $\overbarcal{\mathcal{T}}$ and the 2D design parameters $\mathbf{p}$, aims at maintaining a (manually specified) average triangle distortion $T_{\text{{dist}}}$ and surface area $T_{\text{{area}}}$. Notice that these parameters are constant for all garments, therefore we only need to set them once. 
We implement  $\phi()$ based on the method proposed by Narain \textit{et al.}~\cite{narain2012arcsim}.
We write the optimized mesh as $~\overbarcal{\mathcal{M}} = \{{\mathbf{V}}^{~\overbarcal{\mathcal{M}}}, {\mathbf{E}}^{~\overbarcal{\mathcal{M}}}\}$, where ${\mathbf{V}}^{~\overbarcal{\mathcal{M}}} \in \mathbb{R}^{3 \times {{V}^{~\overbarcal{\mathcal{M}}}}}$ are the vertices of the \new{optimized surface, and ${\mathbf{E}}^{~\overbarcal{\mathcal{M}}}$ the edges of the mesh}.
Figure \ref{fig:retopo} shows an example of the template topology $\lowpmean$ for a long dress design, which result in many degenerated triangles, and the optimized topology $\overbarcal{\mathcal{M}}$.
\new{In practice, $\phi()$ works in the UV-space of the 2D panels, which are automatically sew together to obtain  $\overbarcal{\mathcal{M}}$. We have simplified the notation for the sake of clarity.}
Notice that the \textit{surface} of $\overbarcal{\mathcal{M}}$ and $\lowpmean$ is analogous, but their topology is different.

\paragraph*{Shape-Dependent Smooth Garment Deformation.} Having the optimized mesh topology $\overbarcal{\mathcal{M}}$ computed, in this second step we address the modeling of garment deformations caused by the target body shape. 
\new{To represent parametric bodies, we use the popular model SMPL~\cite{loper2015smpl}, which provides a PCA-based representation of human bodies in T-pose, parameterized by $\beta \in \mathbb{R}^{10}$. We use the first component throughout the paper, since it encapsulates the largest variance in body shape. Importantly, SMPL also provides per-vertex rigging weights $\mathbf{w}_i$, which we use later in this section as a descriptor for garment vertices.}

\begin{figure*}
	\centering
	\includegraphics[width=0.95\linewidth]{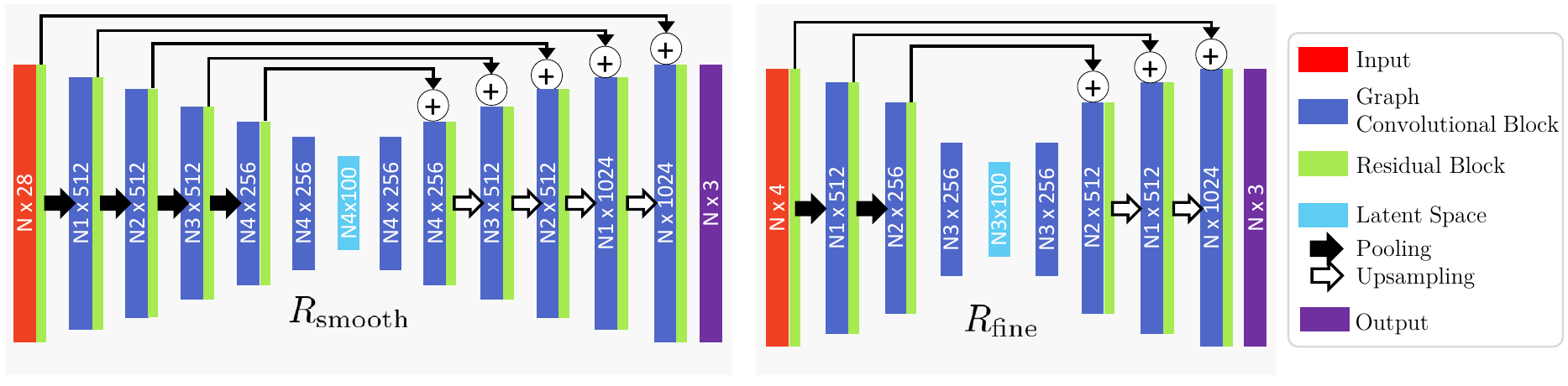}
	\caption{UNet-based architecture for regressors {\normalfont $R_{\text{smooth}}$} (left) and {\normalfont $R_{\text{fine}}$} (right). Each pooling or upsampling pass reduces or augments the number of nodes to half or double size. The input number of nodes is the same for both regressors, they differ in the number of intermediate layers, which is bigger for {\normalfont $R_{\text{smooth}}$} as it has to learn a broader range of deformations.}
	\label{fig:pipeline_unet_smoot}
\end{figure*}
We therefore seek to learn a regressor $R_{\text{smooth}}$
that deforms the mean shape garment $\overbarcal{\mathcal{M}}$ and outputs a mesh that reproduces a smoothed drape of the garment onto the target body shape $\beta$. 
We design $R_{\text{smooth}}$ such that it learns global and smooth deformations, which has two main advantages: first, it eases the learning task since it reduces the variance in data
and second, it decouples target body-dependent deformations (\textit{i.e.,} global stretching and draping effects) from material-dependent (\textit{i.e.,} fine wrinkles) deformations, which we will learn on a subsequent step.
However, formulating such regression task is not trivial: the topology of the input mesh $\overbarcal{\mathcal{M}}$ is unknown at train time since we generate it at run time depending on the design parameters $\mathbf{p}$.
Therefore, we cannot employ a fully connected network, where the input is a fix-size vector corresponding to the number of vertices of the mesh (a strategy commonly used in most of recent learning-based garment deformation methods \cite{wang2019intrisicspace,wang2018multimodalspace,santesteban2019virtualtryon}) and, instead, we propose to use a graph-based fully convolutional architecture.

Two key ingredients are required to design the regressor $R_{\text{smooth}}$ as a graph-based fully convolutional neural network: first, a convolution operator that is able to deal with graph input and, second, an efficient graph pooling operator that is able to coarsen the mesh by clustering together similar vertices.
Specifically for this work, for graph convolutions we use the operator based on truncated Chebyshev polynomial proposed by Defferrard~\textit{et al.} \cite{defferrard2016convolutional}, which has shown to be very efficient given its linear computational complexity and constant learning complexity, like classical convolutional neural networks (\textit{e.g.,} for images or other Euclidean domains). 
For mesh coarsening we use the approach proposed by Ranjan \textit{et al.}~\cite{ranjan2018generating}, which consists of precomputing down- and upsampling matrices using a traditional method for surface simplification by Garland and Heckbert ~\cite{garland1997surface}.

Having the operators defined, we now explain how we design our fully convolutional regressor $R_{\text{smooth}}$. Starting from the mean shape 3D drape mesh $~\overbarcal{\mathcal{M}} = \{{\mathbf{V}}^{~\overbarcal{\mathcal{M}}}, {\mathbf{E}}^{~\overbarcal{\mathcal{M}}}\}$, we first build an analogous undirected graph $~\overbarcalT{\mathcal{G}}=(\mathbf{N}, \mathbf{C})$,
with as many nodes and edges, as vertices and edges in the mesh, $\mathbf{N} = \mathbf{V}^{~\overbarcal{\mathcal{M}}} \in \mathbb{R}^{3 \times {{V}^{~\overbarcal{\mathcal{M}}}}}$ and $\mathbf{C} =  \mathbf{E}^{~\overbarcal{\mathcal{M}}} \in \mathbb{R}^{3 \times {{E}^{~\overbarcal{\mathcal{M}}}}}$, which we wish to use as input to the graph neural network. 
However, using vertices position as a descriptor for the graph nodes does not leverage all the information available in this context.
Our key observation is that we can also append \textit{semantic body part} information into the graph. To this end, for each garment vertex $\mathbf{v}_i^{~\overbarcal{\mathcal{M}}}$
we find the closest body vertex $\mathbf{v}_k^{{\mathcal{B}}}$ , and append its associated rigging weights $\mathbf{w}_k$ into each graph node descriptor.
Additionally, we also append the shape descriptor $\beta$ to each node.
Therefore, the $i^{\text{th}}$ node of the graph $\overbarcalT{\mathcal{G}}$ is defined as $\mathbf{n}_i = \{\mathbf{v}_i^{~\overbarcal{\mathcal{M}}},\mathbf{w}_k,\beta\} \in \mathbb{R}^{3+J+|\beta|}$, where $J$ is the number of body joints (24 for SMPL \cite{loper2015smpl}), and $|\beta|$ the number of shape coefficients (1 for the results shown in this paper).

We then input the graph $\overbarcalT{\mathcal{G}}$ into our fully convolutional regressor
\begin{equation}
R_{\text{smooth}}(\overbarcalT{\mathcal{G}}) = \Delta_{\text{smooth}}
\end{equation}
to predict a vector of  3D displacements $\Delta_{\text{smooth}} \in \mathbb{R}^{3 \times {{V}^{~\overbarcal{\mathcal{M}}}}}$. 
The architecture of the network, inspired by the success of fully convolutional U-Net~\cite{ronneberger2015u} for image segmentation, is depicted in Figure~\ref{fig:pipeline_unet_smoot}.
The final deformed mesh of this second stage is then computed by adding the predicted 3D offsets to the mean shape 3D drape
\begin{equation}
\highpbetasmooth = ~\overbarcal{\mathcal{M}} + \Delta_{\text{smooth}}.
\end{equation}  
To train the regressor $R_{\text{smooth}}$
we create a dataset of ground-truth deformations of two different materials and a range of body shapes using the physics-based cloth simulation \cite{narain2012arcsim}. 
We leverage the whole set of training data 
without introducing bias due to material-dependent deformations by first applying a Laplacian smoothing operator to each generated mesh, and then computing the average of each corresponding sample (\textit{i.e.,} those with same topology, garment type, and target shape) before substracting it from the mean shape to obtain the displacements $\Delta_{\text{smooth}}^{\text{GT}}$.
As a loss function we use the $\ell^2$-norm of the error between ground truth displacements and predictions, in addition to the $\ell^2$ regularization of the network weights

\subsection{Fine 3D Body Drape}
\label{sec:accurate_drape}
The garment mesh $\highpbetasmooth$ successfully reproduces the global garment deformations due to target body shape, but lacks fine details that depend largely on the material. We address such source of deformations in this third and last step by further deforming the garment mesh. 
To this end, we learn to regress a new set of 3D displacements $\Delta_{\text{fine}}$ using a fully convolutional network that takes as input a graph $\mathcal{G}$ built from the vertices positions $ \mathbf{v}_i^{{\mathcal{M}_{\text{smooth}}}}$ and its associated rigging weights, analogous to the graph $\overbarcalT{\mathcal{G}}$ described in Section \ref{sec:smooth_drape}
\begin{equation}
R_{\text{fine}}(\mathcal{G}) = \Delta_{\text{fine}}.
\end{equation}
Our final predicted 3D drape $\highpbetafine$ is then computed by adding the fine displacements onto the mesh  $\highpbetasmooth$
\begin{equation}
\highpbetafine = \highpbetasmooth + \Delta_{\text{fine}}.
\end{equation} 
To train the regressor $R_{\text{fine}}$ we use the same simulated fits as in Section \ref{sec:smooth_drape}. However, in this case, we take advantage of the material-dependent deformations and train one regressor per material type. We
generate the ground truth offsets  $\Delta_{\text{fine},~m}^{\text{GT}}$ per each material $m$ by substracting the smoothed fits from the simulated fits.
As loss function for $R_{\text{fine}}$ we use the same loss as $R_{\text{smooth}}$, with the ground truth fine-scale displacements $\Delta_{\text{fine},~m}^{\text{GT}}$ instead. 
	\subsection{Self-Supervised Learning of Body-Garment Collisions}
\label{sec:selfsupervised_collisions}
The objective losses used to train regressors $R_{\text{smooth}}$
and $R_{\text{fine}}$ minimize the reconstruction error but, due to expected residual errors in unseen shapes and topologies, this term alone does not guarantee predicted deformations to be free of body-garment collisions.
This is a common issue in learning based solutions, which has been address with rendering tricks \cite{de2010stable}, postprocessing steps \cite{santesteban2019virtualtryon}, or explicit collision loss terms \cite{gundogdu2019garnet} using supervised training.
Inspired by the later, we propose a collision loss term that we can train in a \textit{self-supervised strategy}, and therefore does not require to generate expensive ground truth simulations. This is a major advantage over previous explicit collision losses.

Specifically, for each vertex of the garment
 $\mathbf{v}_i^{~\mathcal{M}}$
we find the closest body vertex $\mathbf{v}_k^{{\mathcal{B}}}$ and compute the collision loss as 
\begin{equation}
\mathcal{L}_\text{collision} = max(-\mathbf{n}_k^{{\mathcal{B}}}(\mathbf{v}_i^{~\mathcal{M}} - \mathbf{v}_k^{{\mathcal{B}}}), 0),
\label{eq:collision}
\end{equation}
where $\mathbf{n}_k^{{\mathcal{B}}}$ is the normal vector of the body vertex. The work of Gundogdu \textit{et al.}~\cite{gundogdu2019garnet} uses this loss to penalize collisions during training, but unless the train dataset is exhaustive enough, this approach does not guarantee collision-free results for unseen inputs. 
In our particular case this is particularly bad, since creating an exhaustive dataset of cloth simulations is not feasible due to the arbitrary topology input of our method.

\new{Therefore, starting from network weights trained for $R_{\text{smooth}}$ and $R_{\text{fine}}$,} we propose a novel strategy to fine-tune our networks using Equation \ref{eq:collision} to produce collision-free results for arbitrary inputs. The key insight of our approach is that evaluating the collision loss does \textit{not} require ground-truth data. Therefore, we can feed the network with random inputs and train on the collision loss \new{only} until it converges to a value near zero.
To this end, during the self-supervised step we sample random body shapes $\beta$ and garment topologies $\overline{\mathcal{M}}$, feed them into our pipeline, and use the predicted mesh to fine-tune $R_{\text{fine}}$ with Equation \ref{eq:collision}. Thanks to this strategy the number of collisions has been reduced by 70\% during training, and 20\% in validation.
	
\section{Evaluation and Results}
\label{sec:results_and_evaluation}
In this section we quantitatively and qualitatively evaluate our results in different scenarios.
Specifically, we demonstrate our generalization capabilities, compare with the state-of-the-art method of Santesteban \textit{et al.} \cite{santesteban2019virtualtryon}, and with a newly proposed brute force baseline for parametric virtual try-on.

\paragraph*{Dataset and Implementation Details.} %
Our ground truth dataset has been generated from 19 different garment pattern designs, two different topologies per design, and 201 values for the body shape $\beta$ \new{from the SMPL body model\cite{loper2015smpl}}, uniformly sampled within the range -3 and 3 (from which 100 have been exclusively used for test).
The resulting meshes have between 1,414 (for the simpler case) and 3,581 (for long dresses) vertices. $\lowpmean$ has a \new{fixed} size of 403 vertices, value which dynamically change for $\highpmean$ depending on the garment complexity after the topology optimization step.
\new{To generate our data for the first step described in Section \ref{sec:mean_drape}, in order to avoid potential topology-related problems (\textit{e.g.}, highly distorted triangles, irregular vertex positions, etc.) at simulation time, we first use a high-resolution mesh of 17,246 vertices, and then consistently downsample the simulated meshes to 403 vertices. 
2D panel meshes are manually generated on a 3D modeling software, and the design parameters interpolate between these hand-made panels.}

We have implemented our pipeline in TensorFlow for an efficient GPU training and execution. The parametric 3D draping is a fully connected layer with 3 input neurons (one per design parameter) and a single hidden layer (of ten neurons) trained for less than a minute. Training the fully convolutional networks $R_{\text{smooth}}$, and $R_{\text{fine}}$ took approximately 20, and 14 hours respectively. Fine-tuning the self-supervised collisions took around one day. Everything was executed on a NVIDIA Titan X with 12GB.
\begin{figure}
	\includegraphics[width=\columnwidth]{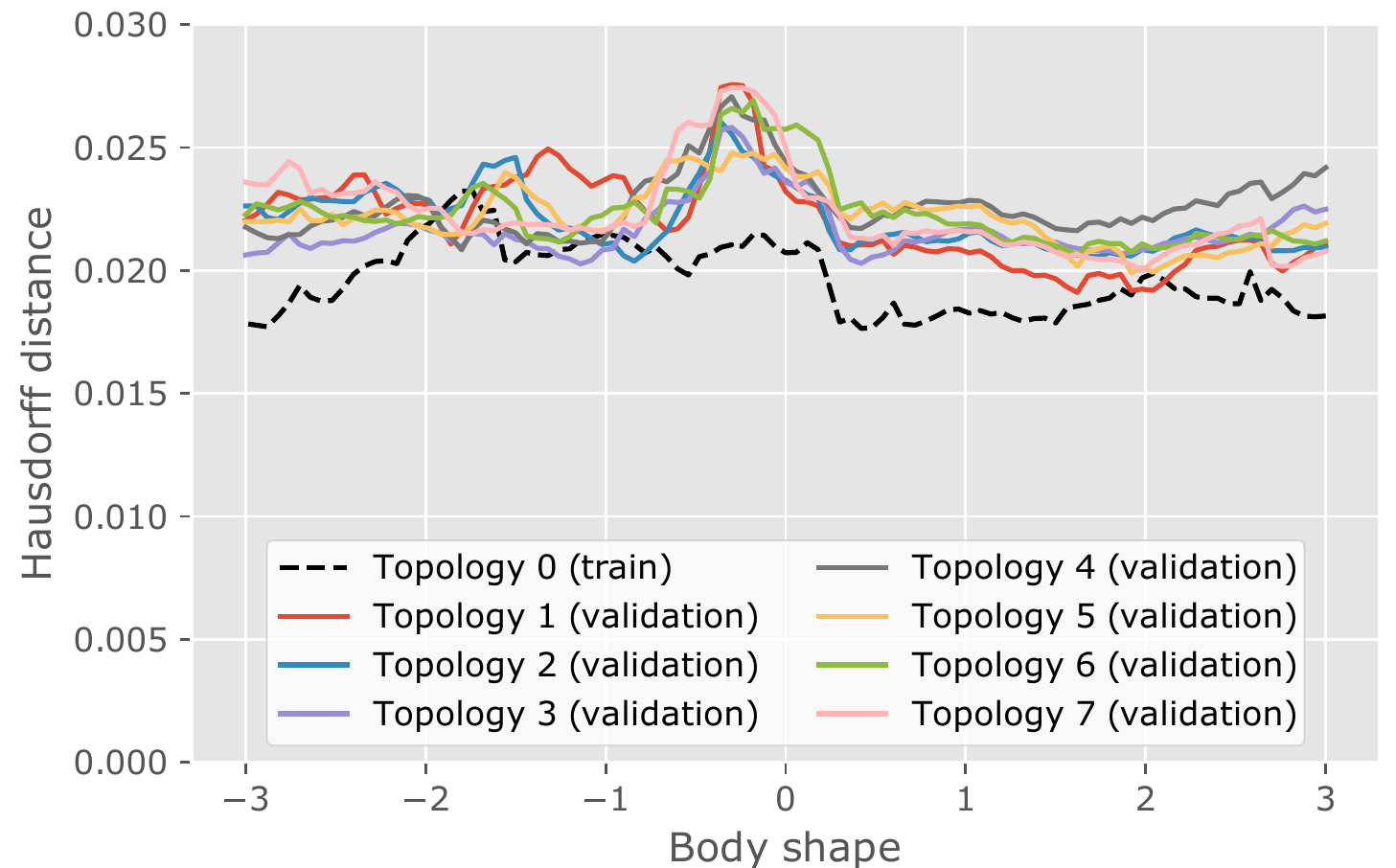}
	\caption{Generalization to new topologies. Hausdorff distance between the predicted and the ground truth meshes for a range of body shapes and 7 validation topologies. 
		Errors in test topologies are consistent, demonstrating the generalization capabilities of our method, and on par to topologies used for training (dashed black).
	}
	\label{fig:new_topologies}
\end{figure}

\paragraph*{Evaluation of Generalization to New Topologies.}
In Figure \ref{fig:new_topologies} we quantitatively evaluate the generalization capabilities of \new{the regressors $R_{\text{smooth}}$ and $R_{\text{fine}}$} to new topologies. 
Specifically, for a given garment parameters $\mathbf{p}$ and material for which we have ground truth simulated data, we randomized the topology (keeping the mean triangle area constant)
of the mean shape predicted mesh $\overbarcal{\mathcal{M}}$, and feed each topology to the regressors $R_{\text{smooth}}$ and $R_{\text{fine}}$ for a range of target shapes $\beta$.
For each predicted mesh, we then compute the Hausdorff distance to the ground truth simulations. Results demonstrate that our method predictions are quantitatively consistent, regardless the topology and target body shape.
Importantly, we also show that the error of the topologies unseen at train time (\textit{i.e.,} validation set) is on par with the error of topologies used to train (in dash black).
\paragraph*{Comparison with Parametric Fully Connected Baseline.}
Despite the lack of methods than can cope with parametric garments due to the need for different topologies, an alternative brute-force approach could be to use a highly-dense topology in $\highpmean$ to represent \textit{all} garments, followed by a fully-connected end-to-end network that predicts displacements over such mesh.
This high dense topology would provide an over-discretized mesh which,
although unnecessarily complex for small garments such as a t-shirt, would provide sufficient details for large garments such as dresses, technically enabling the use of fully-connected pipelines~\cite{santesteban2019virtualtryon}.
We implemented such solution, which can be considered a baseline for data-driven parametric garments, and compared it with our fully convolutional approach.

\begin{figure}[t]
	\includegraphics[width=\columnwidth]{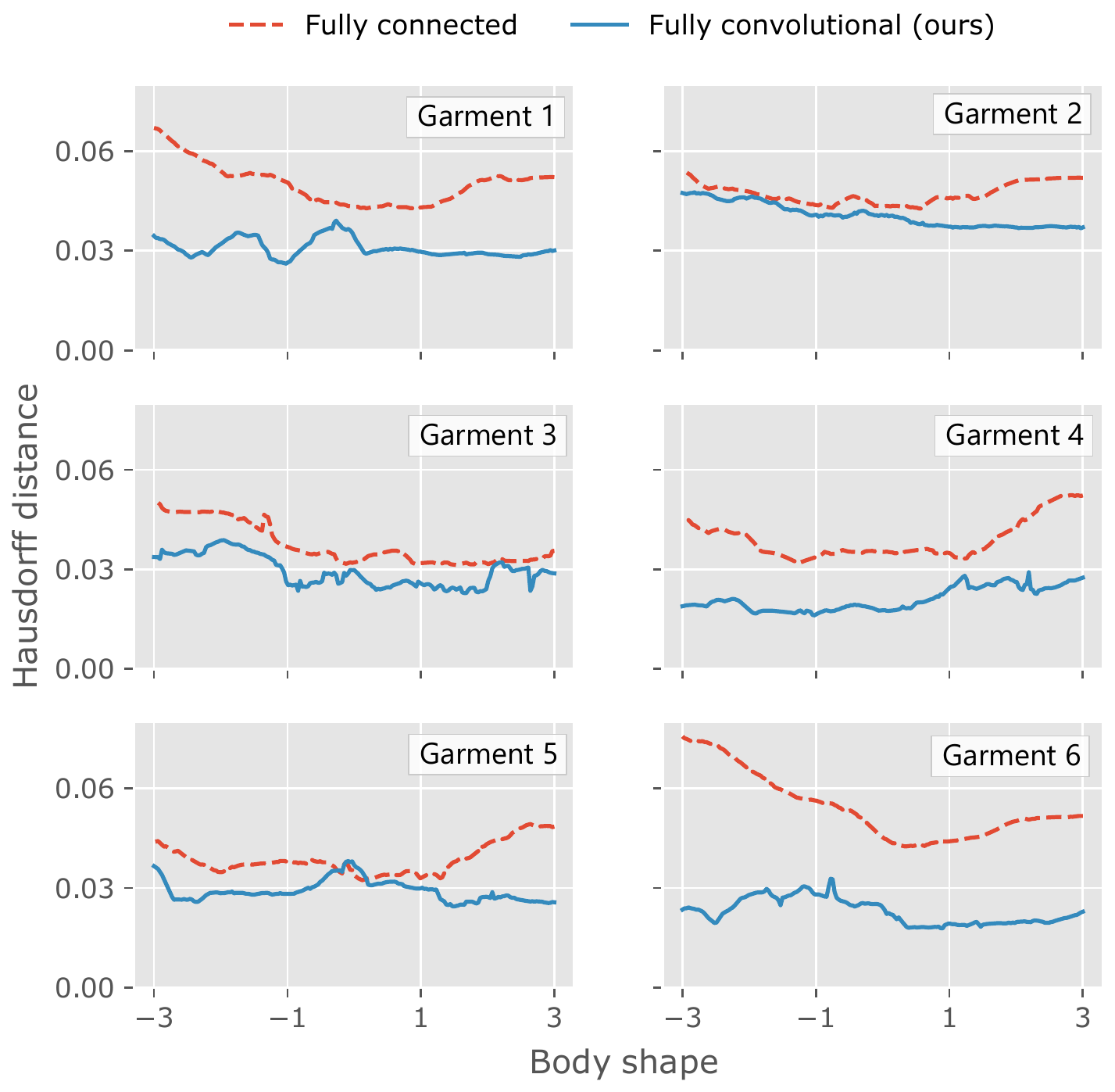}
	\caption{Quantitative evaluation of our fully convolutional (solid blue) approach and the fully-connected baseline (\textit{i.e.}, using the same highly-dense topology for all garments and a fully-connected architecture, dashed red), for 6 garment designs not present in the training set. Our approach consistently outperforms the fully connected baseline since the latter cannot generalize well to unseen garment types.}
	\label{fig:vs_fullyconnected}
\end{figure}

In Figure \ref{fig:vs_fullyconnected} we present a quantitative evaluation of the precision accuracy of our method, and the fully connected baseline. Specifically, for a given garment design (unseen at training time) we compute the Hausdorff error for a range of target body shapes, and demonstrate that our predictions $\highpbetafine$ are consistently more accurate.
Our hypothesis is that the fully-connected approach cannot generalize to garment types outside the training set due to the \textit{global} nature of the densely connected neurons, that are unable to learn local features. 
In contrast, the convolutional nature of our approach is able to capture local features, and therefore correctly predicts deformations of garment types unseen at train time but locally present in train examples.

\begin{figure*}
	\centering
	\includegraphics[width=0.93\linewidth]{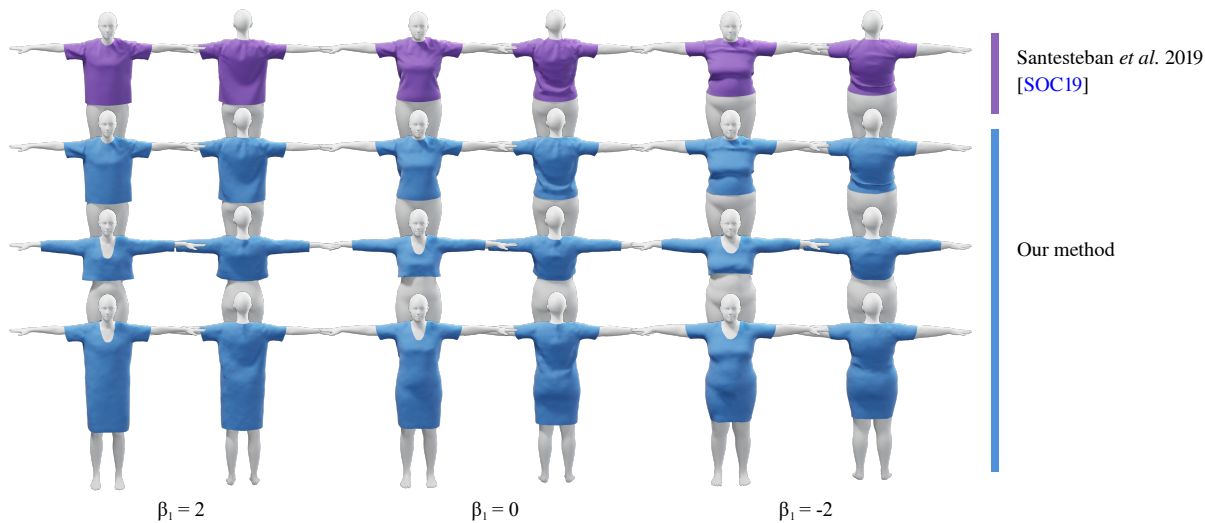}
	\caption{Qualitative comparison with the single-garment and fix topology method of Santesteban \textit{et al.} \cite{santesteban2019virtualtryon} and ours. When sampling the same garment type use to train their method, our results are on par with Santesteban's (rows 1, 2), while our approach allows for a much richer space of garment types and topologies (rows 3, 4).}
	\label{fig:vs_santesteban}
\end{figure*}
\begin{figure}
	\centering
	\begin{subfigure}{.11\textwidth}
		\centering
		\includegraphics[width=\linewidth,trim={600pt 0pt 600pt 100pt},clip]{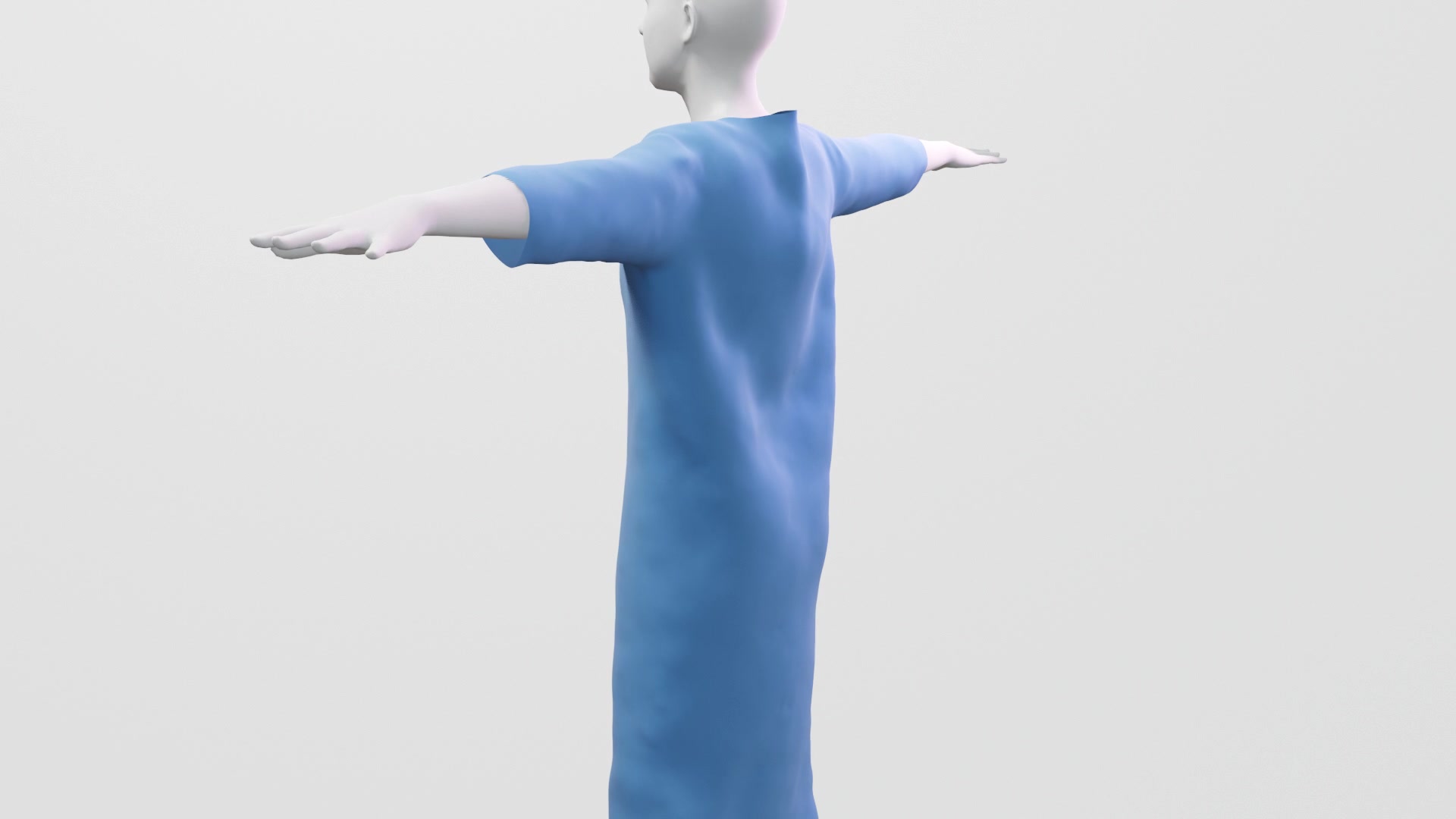}
	\end{subfigure}
	\begin{subfigure}{.11\textwidth}
		\centering
		\includegraphics[width=\linewidth,trim={600pt 0pt 600pt 100pt},clip]{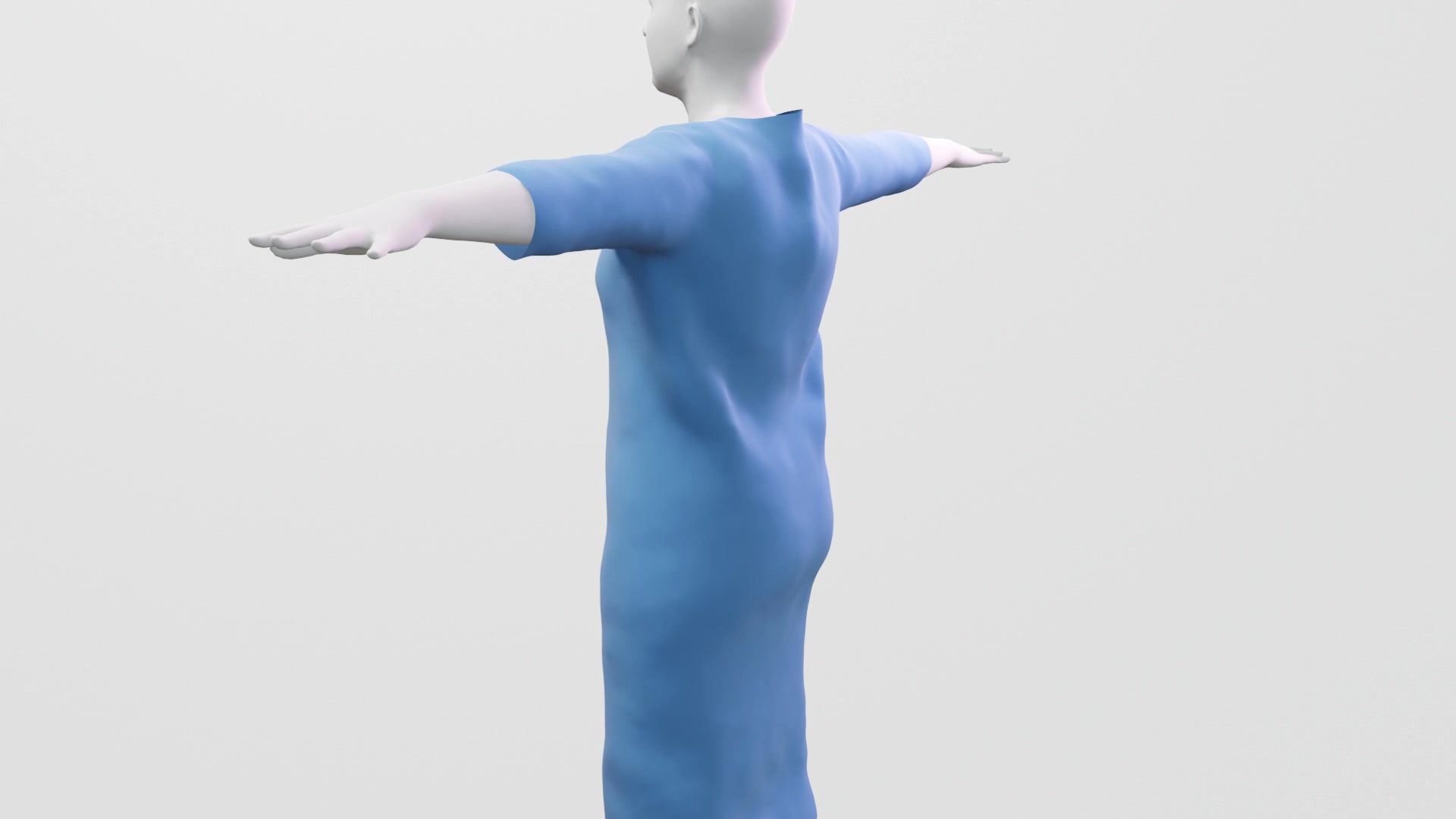}
	\end{subfigure}
	\begin{subfigure}{.11\textwidth}
		\centering
		\includegraphics[width=\linewidth,trim={600pt 0pt 600pt 100pt},clip]{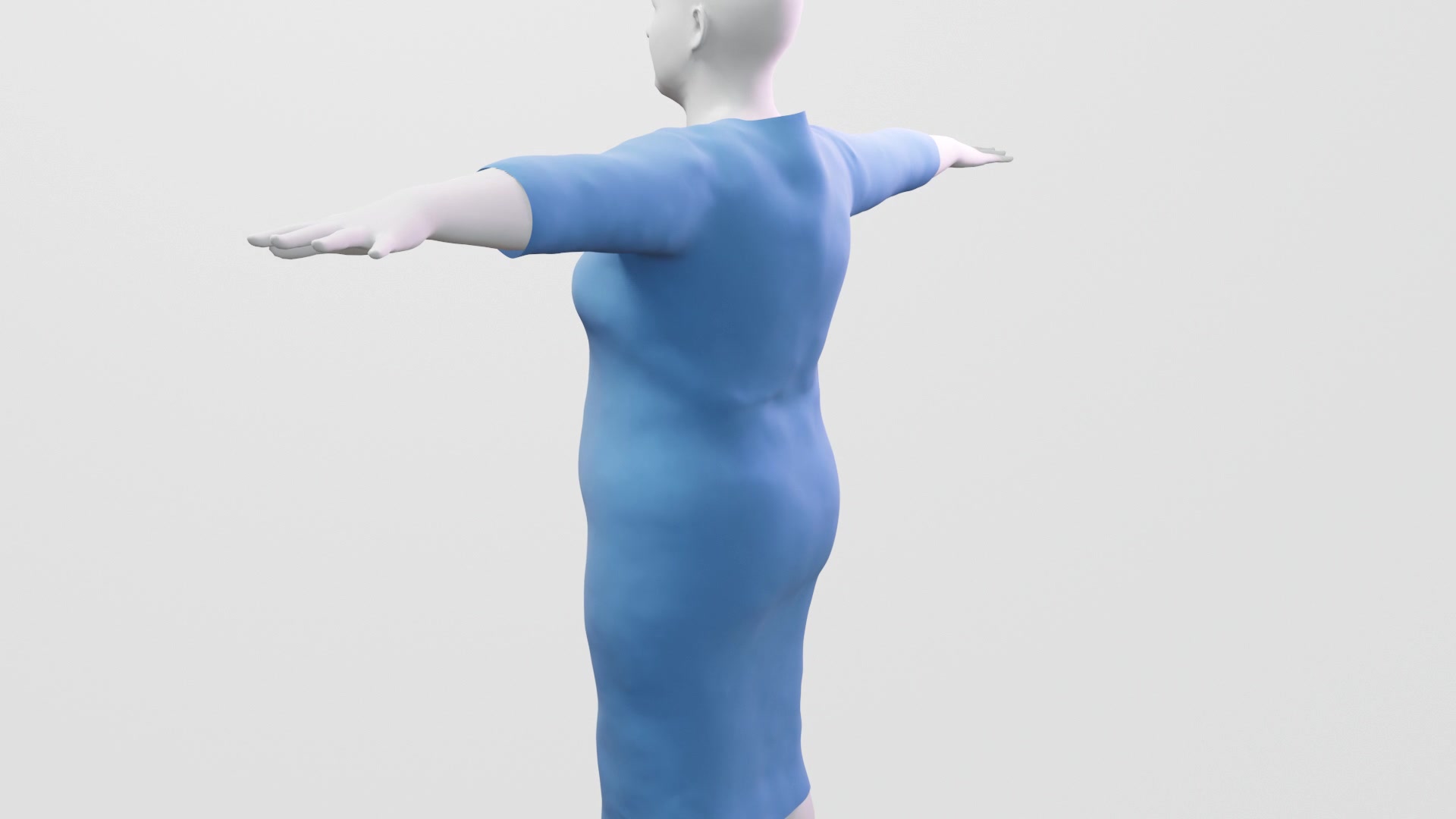}
	\end{subfigure}
	\begin{subfigure}{.11\textwidth}
		\centering
		\includegraphics[width=\linewidth,trim={600pt 0pt 600pt 100pt},clip]{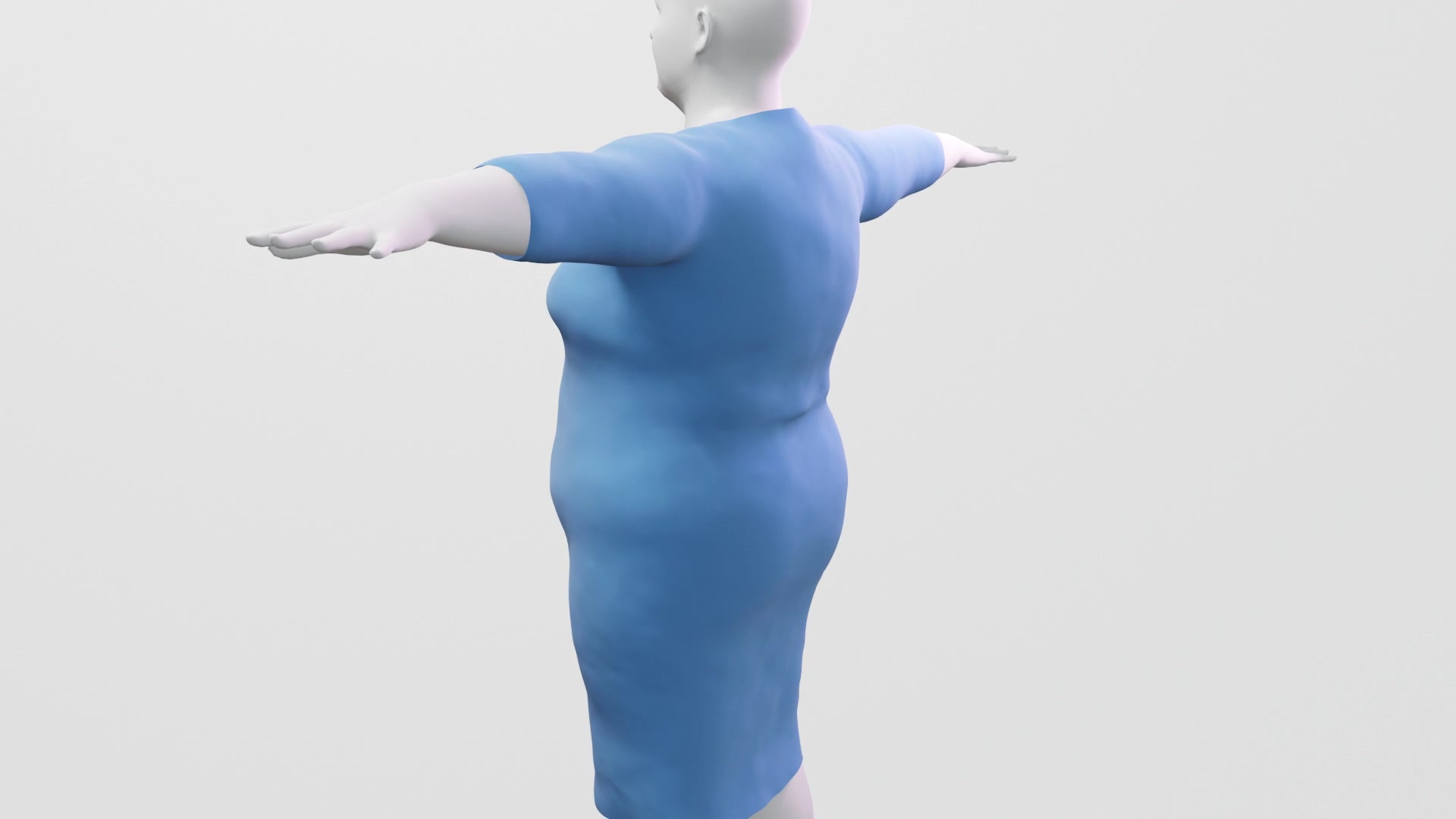}
	\end{subfigure}
	\\[1pt]
	\begin{subfigure}{.11\textwidth}
		\centering
		\includegraphics[width=\linewidth,trim={600pt 0pt 600pt 100pt},clip]{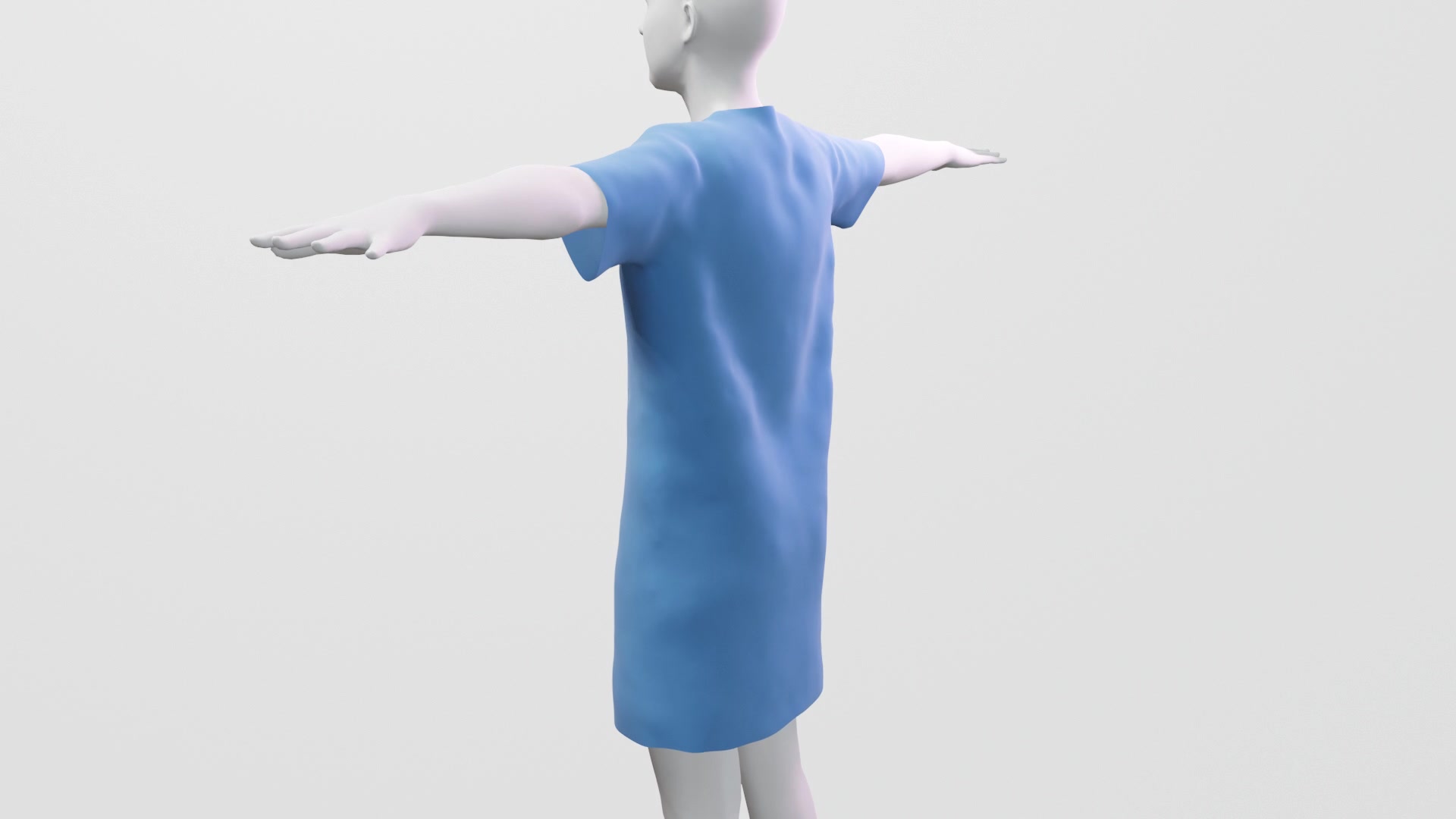}  
	\end{subfigure}
	\begin{subfigure}{.11\textwidth}
		\centering
		\includegraphics[width=\linewidth,trim={600pt 0pt 600pt 100pt},clip]{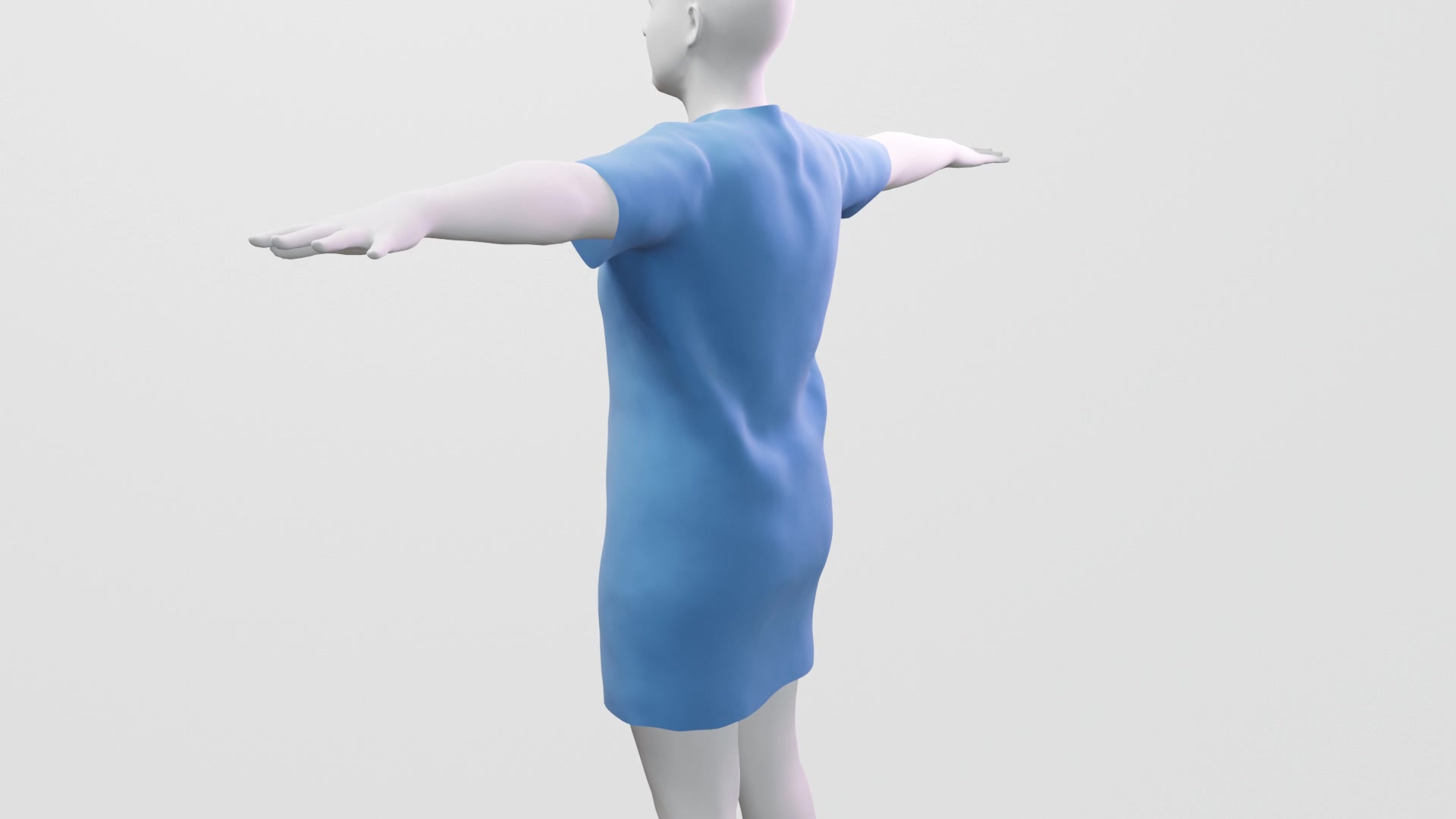}
	\end{subfigure}
	\begin{subfigure}{.11\textwidth}
		\centering
		\includegraphics[width=\linewidth,trim={600pt 0pt 600pt 100pt},clip]{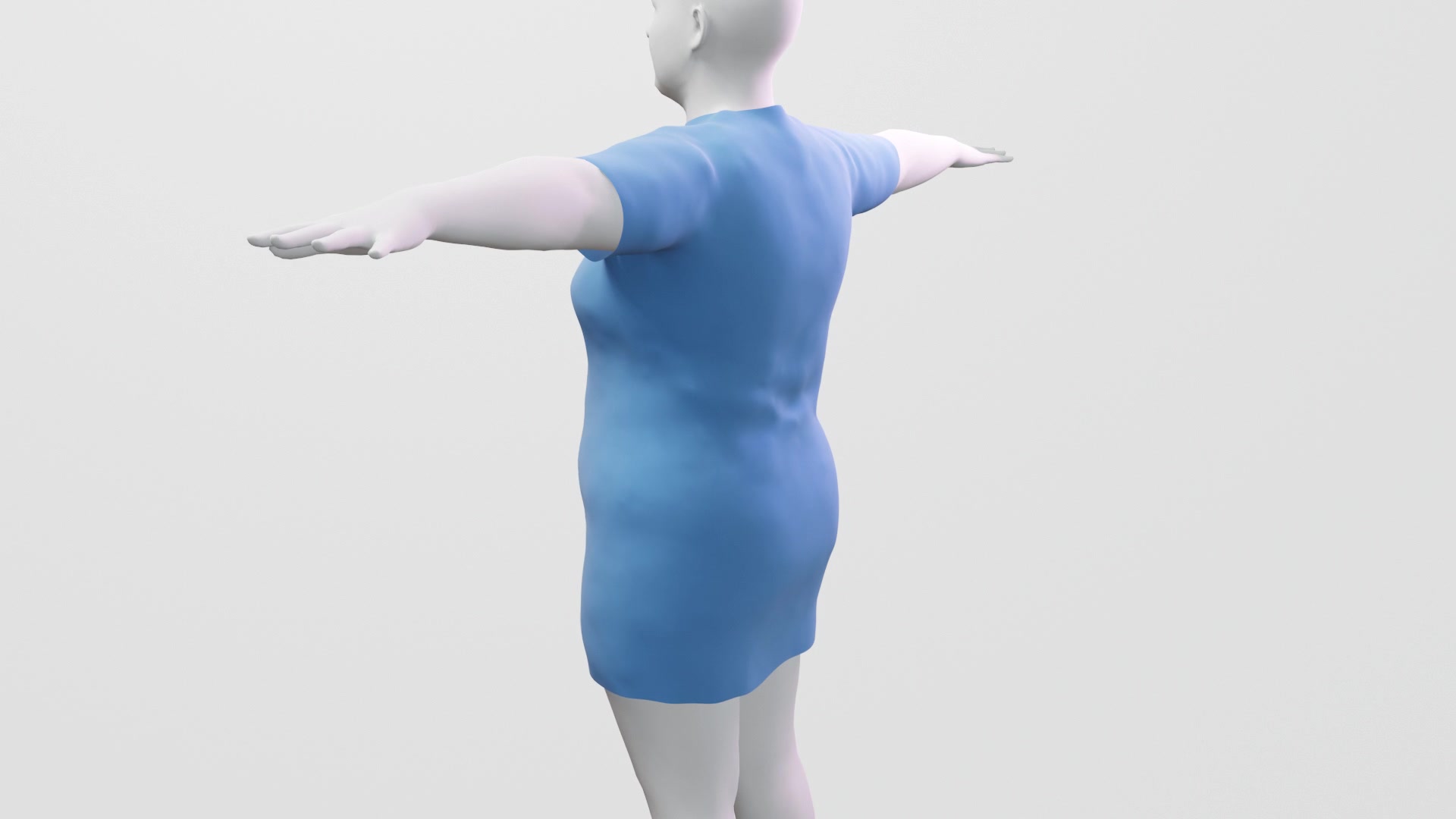}
	\end{subfigure}
	\begin{subfigure}{.11\textwidth}
		\centering
		\includegraphics[width=\linewidth,trim={600pt 0pt 600pt 100pt},clip]{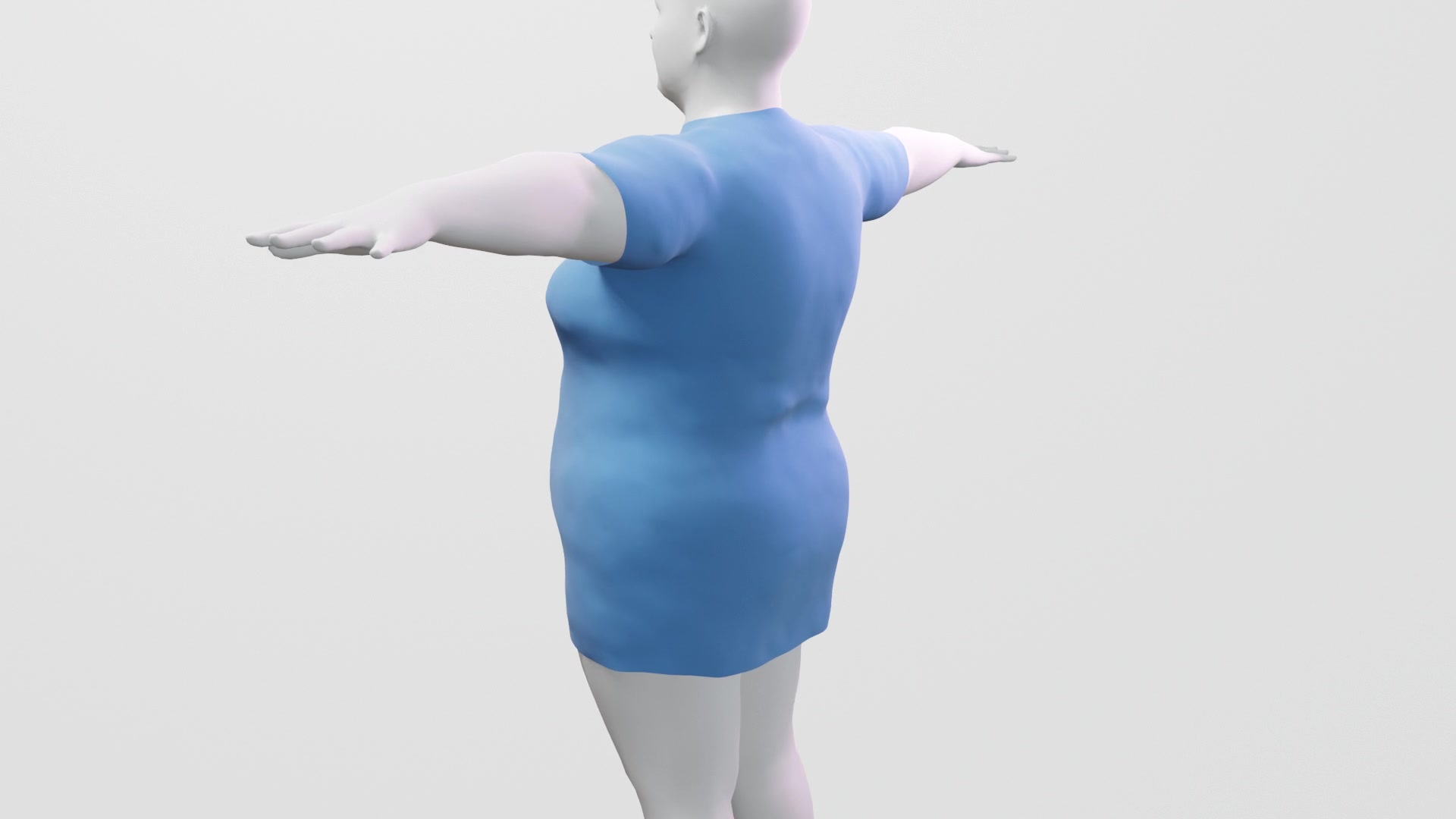}
	\end{subfigure}
	\\[1pt]
	\begin{subfigure}{.11\textwidth}
		\centering
		\includegraphics[width=\linewidth,trim={600pt 0pt 600pt 100pt},clip]{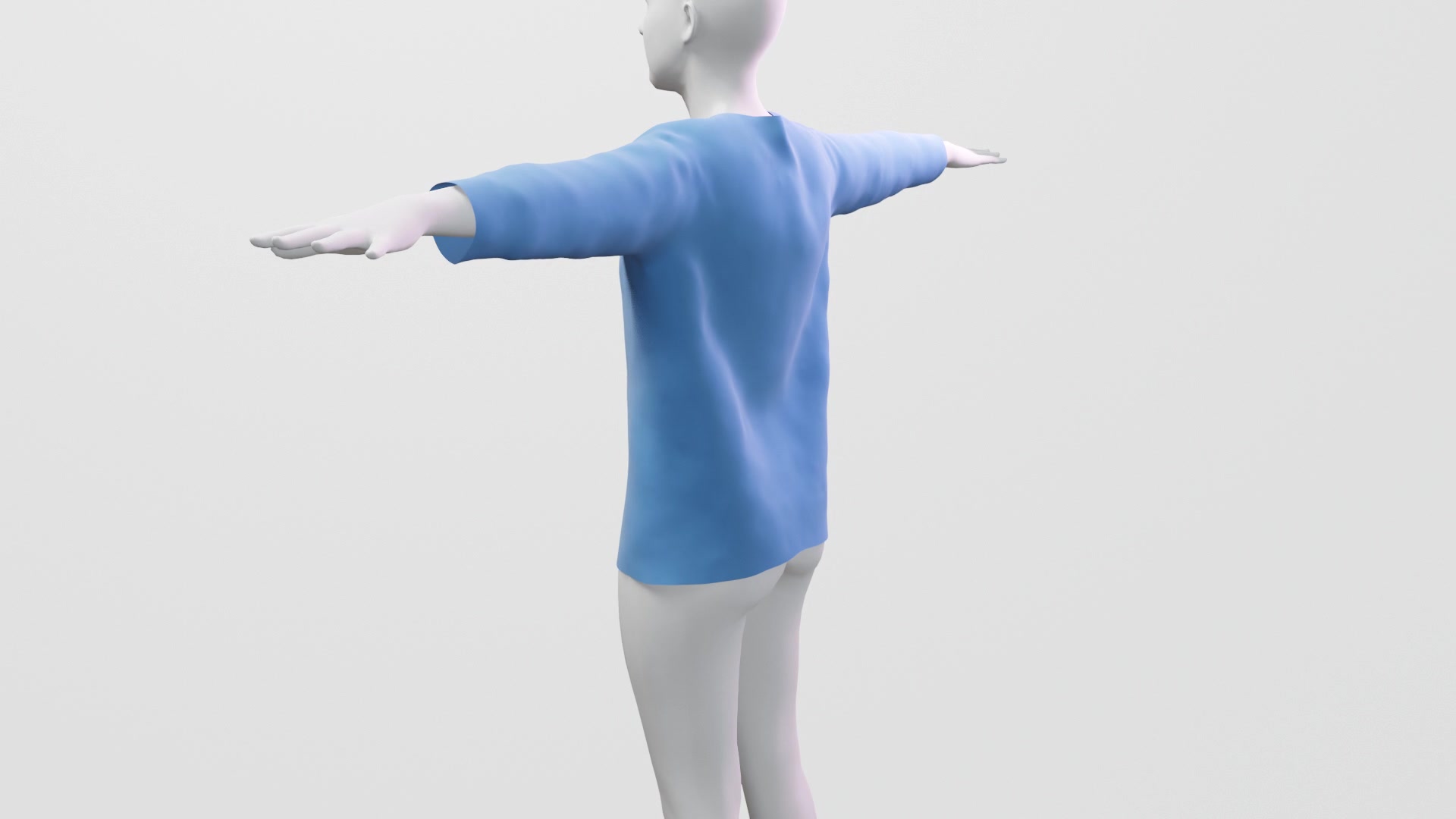}
	\end{subfigure}
	\begin{subfigure}{.11\textwidth}
		\centering
		\includegraphics[width=\linewidth,trim={600pt 0pt 600pt 100pt},clip]{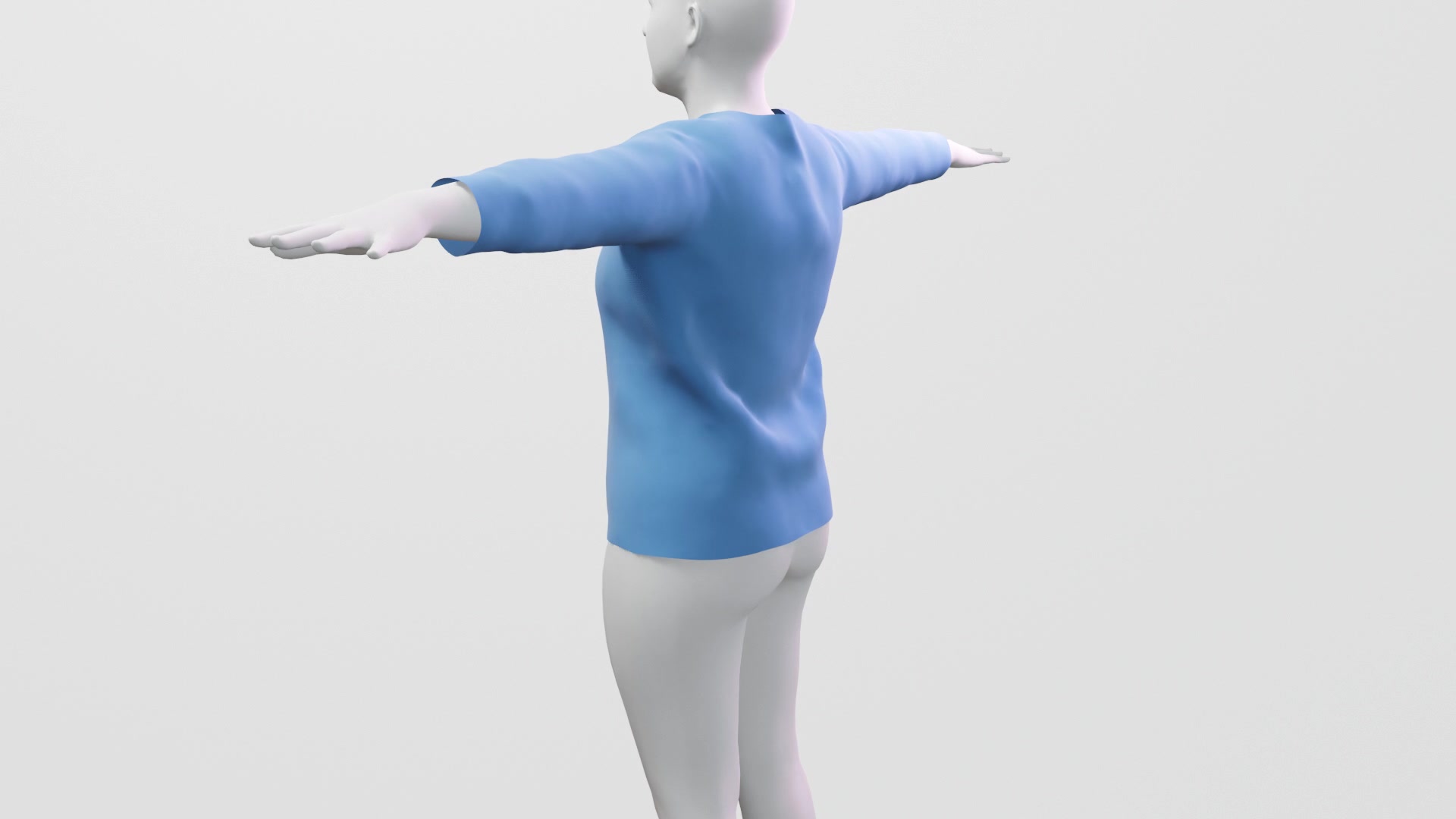}
	\end{subfigure}
	\begin{subfigure}{.11\textwidth}
		\centering
		\includegraphics[width=\linewidth,trim={600pt 0pt 600pt 100pt},clip]{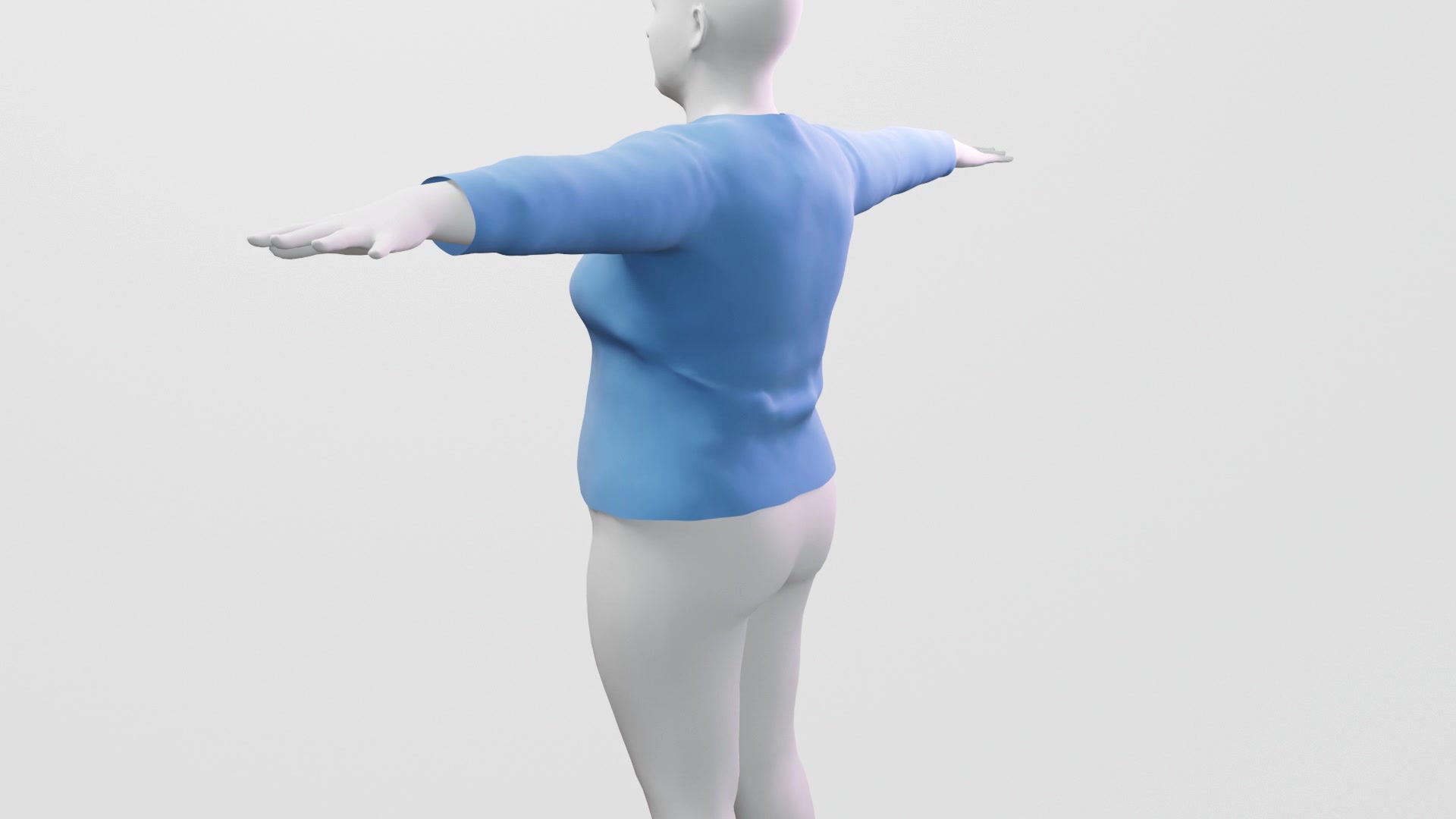}
	\end{subfigure}
	\begin{subfigure}{.11\textwidth}
		\centering
		\includegraphics[width=\linewidth,trim={600pt 0pt 600pt 100pt},clip]{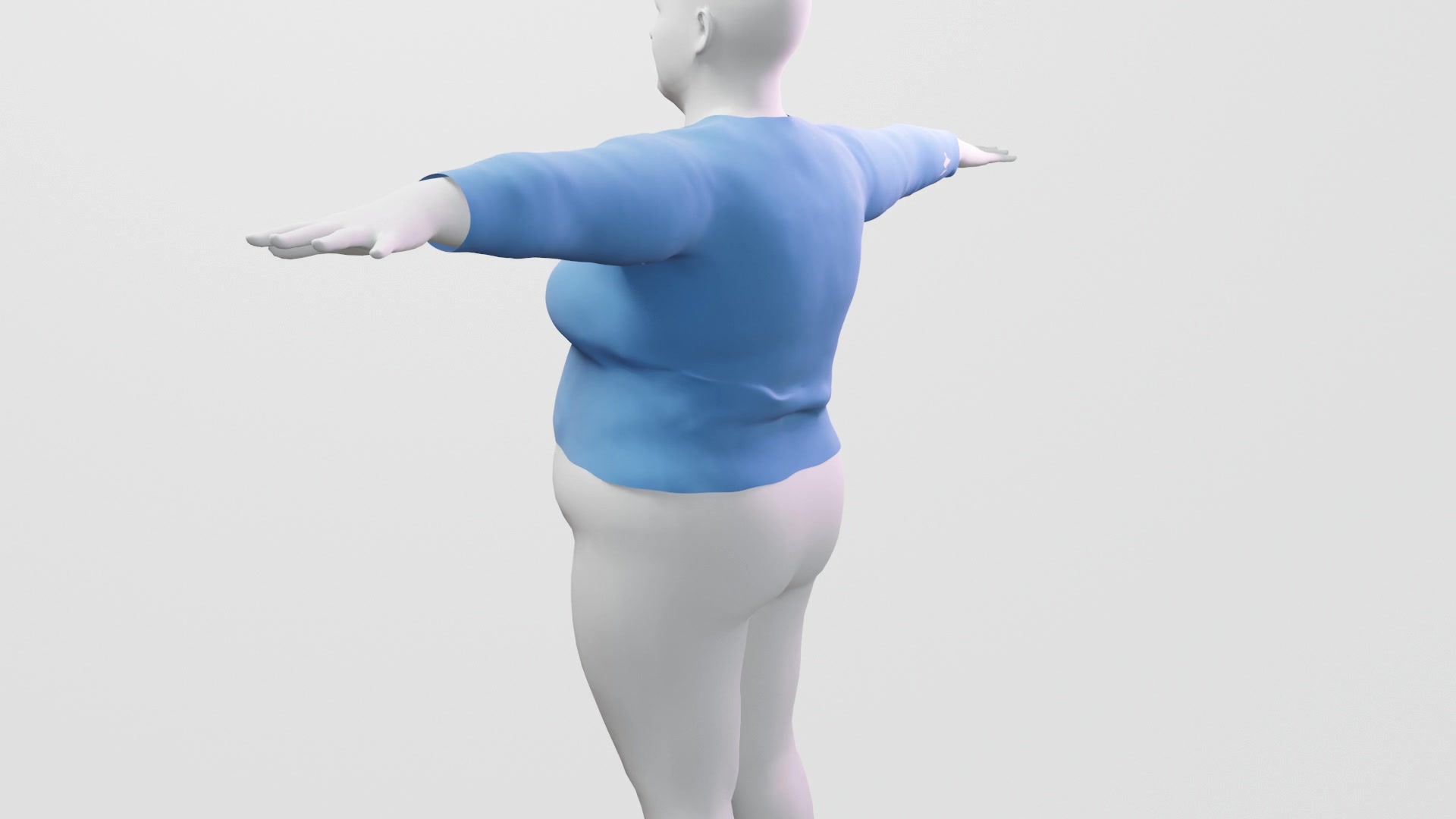}
	\end{subfigure}
	\\[1pt]
	\begin{subfigure}{.11\textwidth}
		\centering
		\includegraphics[width=\linewidth,trim={600pt 0pt 600pt 100pt},clip]{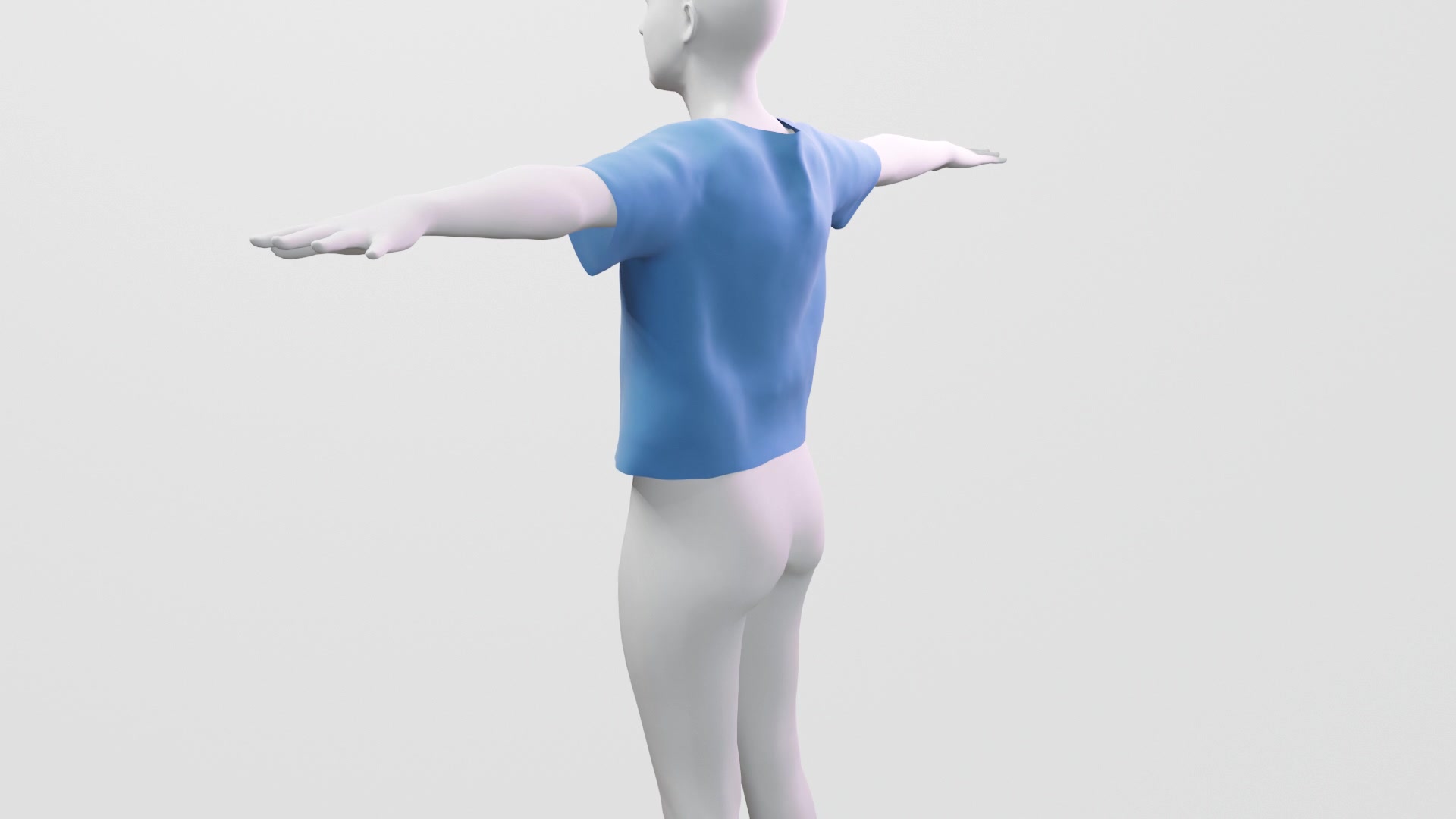}
	\end{subfigure}
	\begin{subfigure}{.11\textwidth}
		\centering
		\includegraphics[width=\linewidth,trim={600pt 0pt 600pt 100pt},clip]{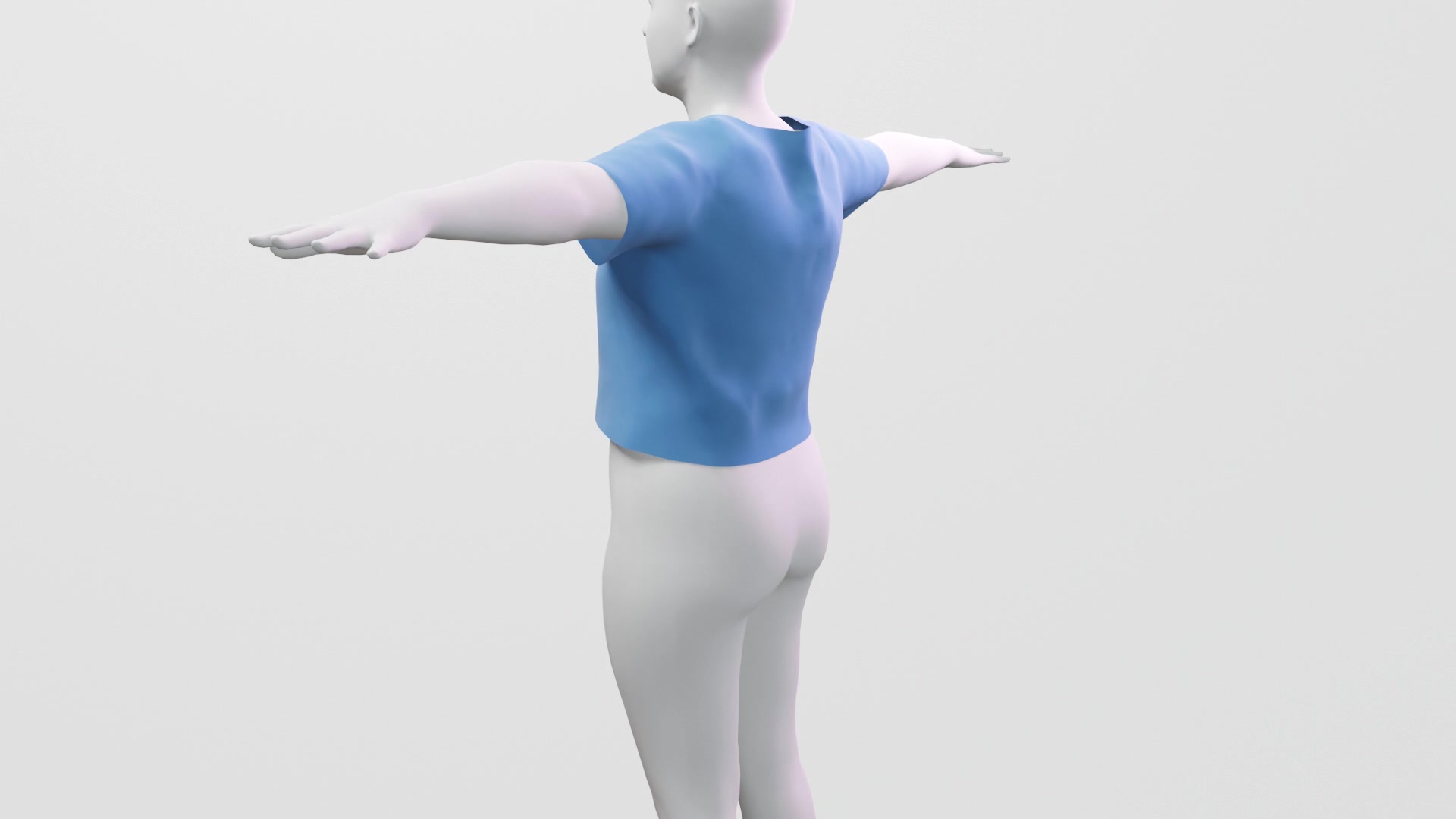}
	\end{subfigure}
	\begin{subfigure}{.11\textwidth}
		\centering
		\includegraphics[width=\linewidth,trim={600pt 0pt 600pt 100pt},clip]{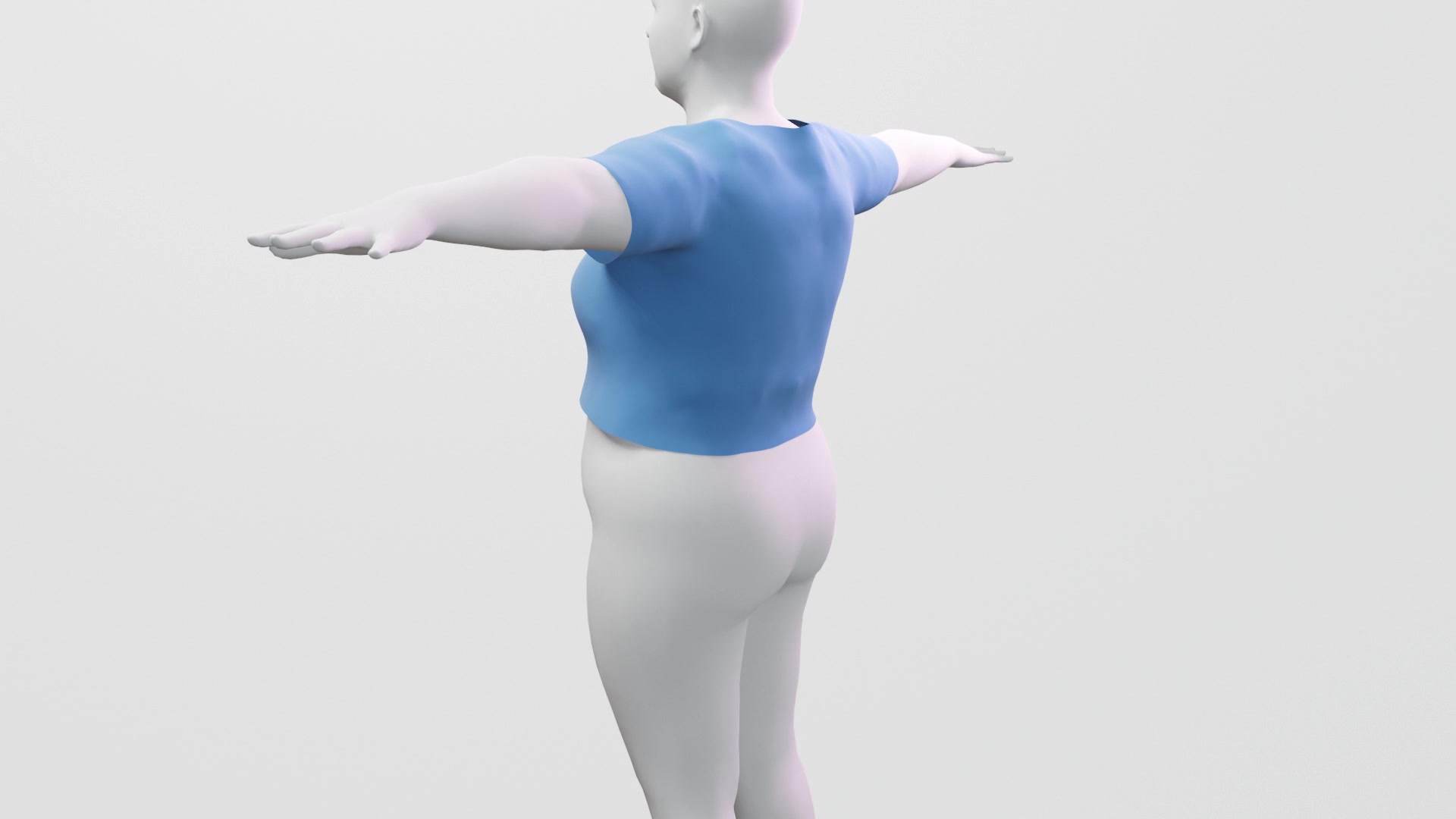}
	\end{subfigure}
	\begin{subfigure}{.11\textwidth}
		\centering
		\includegraphics[width=\linewidth,trim={600pt 0pt 600pt 100pt},clip]{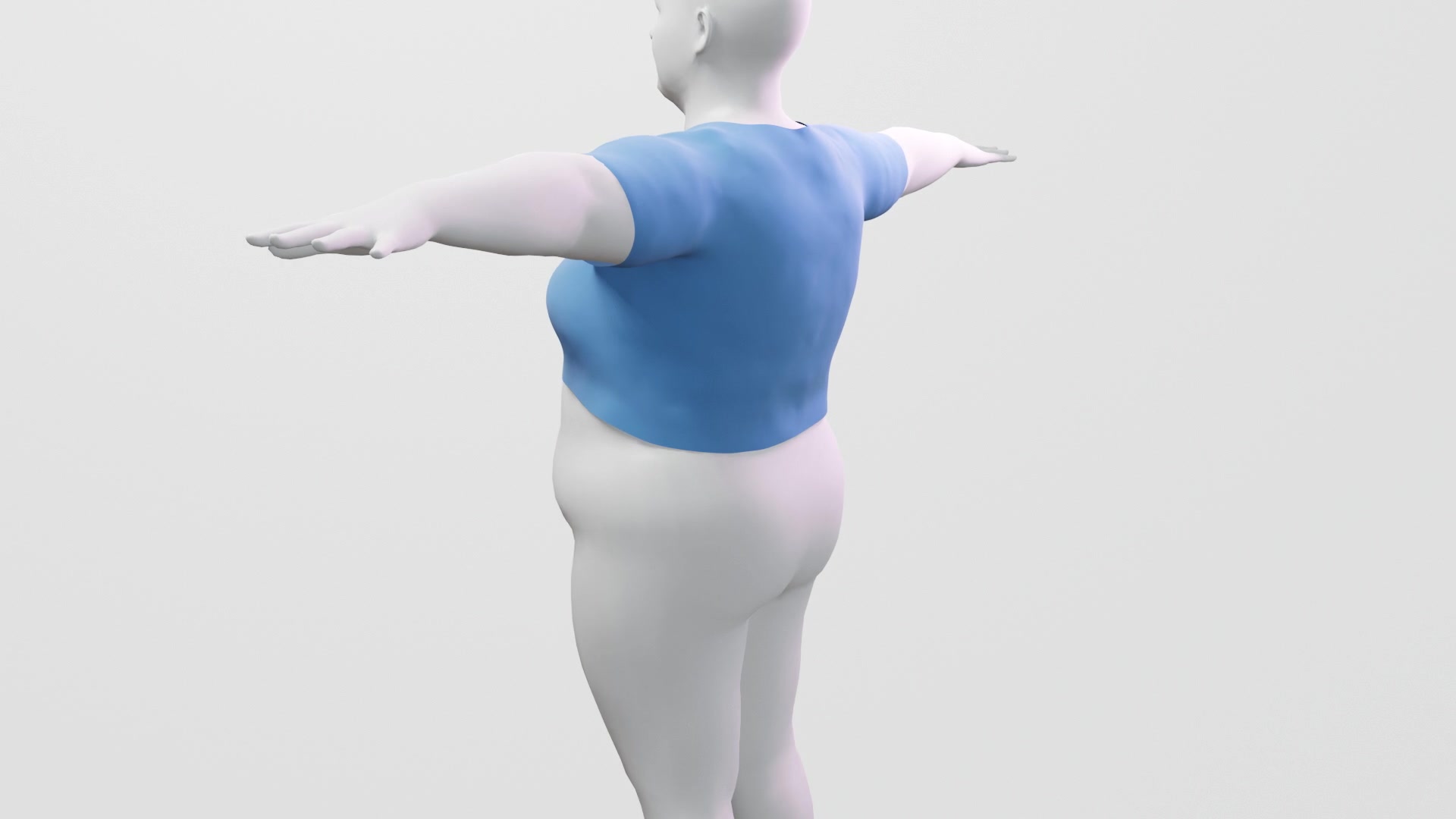}
	\end{subfigure}
	\caption{Virtual try-on results with our method, for a variety of garments (rows), fitted into a range of shapes (columns), both unseen at train time. Our method successfully predicts the drape of the garment, with natural folds and wrinkles at different \new{scales} that depend both on the input garment type and the target body shape.}
	\label{fig:qualitative_evaluation}
\end{figure}
Furthermore, we also evaluate the memory footprint of each method, which also results favorable for us. The fully connected network size is 167 MB, while ours ($R_{\text{smooth}}$ + $R_{\text{fine}}$) is 71MB. This is also expected, since the number of parameters for a fully connected network is significantly higher in comparison to the parameters used in the convolutional kernels. Note also that the fully connected approach needs to be fully trained for any new material while our approach enables easier generalization and transfer learning for new materials through fine-tuning $R_{\text{fine}}$.
\paragraph*{Comparison with Santesteban et al. 2019.}
In Figure \ref{fig:vs_santesteban} we qualitatively compare our results with the state-of-the-art method of Santesteban \textit{et al.}~\cite{santesteban2019virtualtryon}, which is limited to a single garment.
For a garment design analogous to the t-shirt used to train their method, we demonstrate that the predictions of both methods are on par (rows 1 and 2), while we are capable to predict the draping of a much larger number of garments (rows 3 and 4).
This demonstrates the generalization capabilities of our method to arbitrary parametric garment design (and therefore, arbitrary topology).
\paragraph*{Qualitative Results.}
In Figure \ref{fig:qualitative_evaluation} we show qualitative results of our method, for a variety of body shapes, garment types and topologies, all of them \textit{unseen at train time}.
Notice how the wrinkles predicted with our approach naturally match the expected behavior of the garment, and change for each shape-garment pair.
This demonstrates that our method generalizes well to new garment types, topologies, and shapes. 
Check the supplementary video for more qualitative results.

\begin{figure}
	\centering
	\includegraphics[width=0.98\columnwidth]{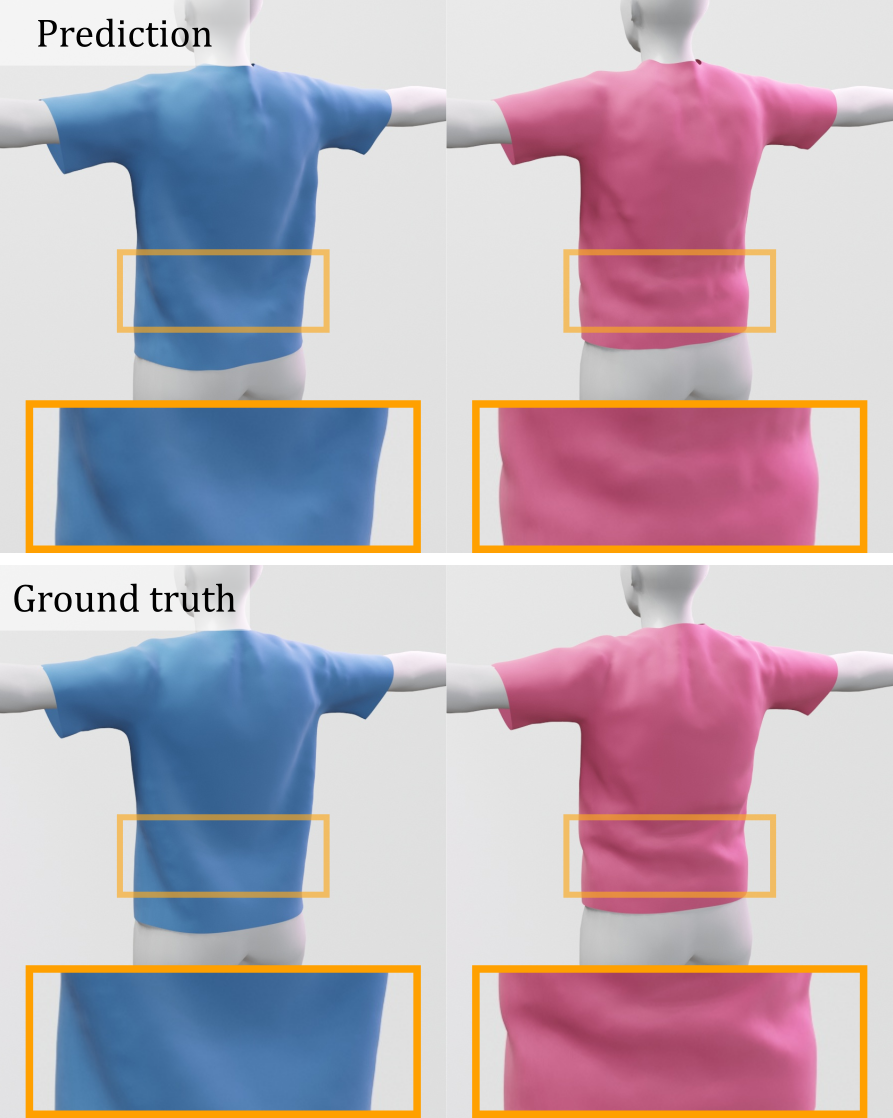}
	\caption{\new{Deformations regressed by our method for two different materials, presented in blue and pink. We demonstrate that, given the same target shape and input garment type, our method (top) is able to learn material-specific details that produce distinctive folds and wrinkles, closely matching the ground truth deformations (down)}.}
	\label{fig:qualitative_materials}
\end{figure}

In Figure \ref{fig:qualitative_materials} we show qualitative predictions of our method, for two different materials, but the same target body shape and garment type (both unseen at train time). 
We demonstrate how our final step $R_{\text{fine}}$ is able to learn material-specific deformations, resulting in visually different folds and wrinkles.
Specifically for this comparison, the blue t-shirt is train on \texttt{gray-interlock} \new{(60\% Cotton, 40\% Polyester)} material and the pink on \texttt{white-dots-on-black} \new{(100\% Polyester)} from ARCSim materials \cite{narain2012arcsim}. 
\new{See \cite{wang2011data} for additional material details}.

	\section{Conclusions}
We have presented a method to predict the drape of a \new{predefined} parametric space of garments onto an arbitrary target body shape.
To achieve this, we propose a novel fully convolutional graph neural network that, in contrast to existing methods, it is not limited to a single garment or topology.
Our novel pipeline, based on U-Net architecture and efficient graph convolutions, generalizes to unseen \new{mesh} topologies, garment \new{parameters}, and body shapes.
To the best of our knowledge, ours is the first fully convolutional approach for virtual \new{try-on} purposes, which opens the door to more general data-driven cloth animation methods based on geometric deep learning.

Despite our step forward in geometric learning-based solutions for cloth animation, our approach still suffers from the following weaknesses that could be addressed by follow up works.
Pose-dependent and material-dependent \new{\textit{input} parameters} are not considered to our approach, and you need to retrain the model to consider these configurations.
Multi-layer garments and contact with external forces are not considered either.
\new{Additionally, commercial garment design probably requires more than 3 parameters. The analysis of the scalability of the proposed method to a larger garment space remains open for future research}.

\paragraph*{Acknowledgments.}
Igor Santesteban was supported by the Predoctoral Training Programme of the Department of Education of the
Basque Government (PRE\_2019\_2\_0104), and Elena Garces was
supported by a Torres Quevedo Fellowship (PTQ2018-009868). The work was also funded in part by the Spanish Ministry of Science (project RTI2018-098694-B-I00
VizLearning).
	\bibliographystyle{eg-alpha-doi} 
	\bibliography{vidaurre_SCA2020}

\end{document}